\documentclass[journal]{IEEEtran}

\usepackage{amssymb}
\usepackage{cite}
\usepackage[pdftex]{graphicx}
\usepackage{epstopdf}
\usepackage[cmex10]{amsmath}
\usepackage{algorithmic}
\usepackage{array}
\usepackage[caption=false,font=footnotesize]{subfig}
\usepackage{enumerate}
\usepackage{enumitem}
\usepackage{bbm}
\usepackage{url}
\usepackage[named]{algo}
\usepackage{multirow}
\usepackage{hhline} 
\usepackage{ragged2e}

\def\btheta{{\boldsymbol \theta}}
\def\bmu{{\boldsymbol \mu}}
\def\bq{{\mathbf q}}
\def\bx{{\mathbf x}}
\def\bc{{\mathbf c}}
\def\bu{{\mathbf u}}
\def\bS{{\mathbf S}}
\def\mS{{\mathcal S}}
\def\bbM{{\mathbb M}}
\def\bbMI{{\mathbb{MI}}}
\def\bbL{{\mathbb L}}
\def\bR{{\mathbf R}}
\def\bI{{\mathbf I}}
\def\bSigma{{\boldsymbol \Sigma}}

\def\imgwidth{0.30\textwidth}
\def\imgheight{0.30\textwidth}
\def\imgspacing{0.01\textwidth}

\begin{document}

\title{Multimodal MRI Neuroimaging with Motion Compensation Based on Particle Filtering$^*$}

% Affiliations
% \author{
% \IEEEauthorblockN{Yu-Hui Chen\IEEEauthorrefmark{1},
% Roni Mittelman\IEEEauthorrefmark{1},
% Boklye Kim\IEEEauthorrefmark{1},and
% Alfred Hero\IEEEauthorrefmark{1}} \\
% \IEEEauthorblockA{\IEEEauthorrefmark{1}University of Michigan, Ann Arbor, MI USA} 
% }

\author{
\IEEEauthorblockN{Yu-Hui Chen$^\dag$,
Roni Mittelman$^\dag$,
Boklye Kim,
Charles Meyer, and
Alfred Hero} \\
\IEEEauthorblockA{University of Michigan, Ann Arbor, MI USA} 
}

% % The paper headers
% \markboth{Journal of \LaTeX\ Class Files,~Vol.~13, No.~9, September~2014}%
% {Shell \MakeLowercase{\textit{et al.}}: Bare Demo of IEEEtran.cls for Journals}

% The only time the second header will appear is for the odd numbered pages
% after the title page when using the twoside option.

% If you want to put a publisher's ID mark on the page you can do it like
% this:
%\IEEEpubid{0000--0000/00\$00.00~\copyright~2014 IEEE}
% Remember, if you use this you must call \IEEEpubidadjcol in the second
% column for its text to clear the IEEEpubid mark.

% make the title area
\maketitle
\let\thefootnote\relax\footnotetext{$^*$This research was partially supported by ARO MURI grant W911NF-15-1-0479 and NIH grant 2P01CA087634-06A2.}
\let\thefootnote\relax\footnotetext{$^\dag$The two authors contributed equally to this work.}

% As a general rule, do not put math, special symbols or citations
% in the abstract or keywords.
\begin{abstract}
Head movement during scanning impedes activation detection in fMRI studies. Head motion in fMRI acquired using slice-based Echo Planar Imaging (EPI) can be estimated and compensated by aligning the images onto a reference volume through image registration. However, registering EPI images volume to volume fails to consider head motion between slices, which may lead to severely biased head motion estimates. Slice-to-volume registration can be used to estimate motion parameters for each slice by more accurately representing the image acquisition sequence. However, accurate slice to volume mapping is dependent on the information content of the slices: middle slices are information rich, while edge slides are information poor and more prone to distortion. In this work, we propose a Gaussian particle filter based head motion tracking algorithm to reduce the image misregistration errors. The algorithm uses a dynamic state space model of head motion with an observation equation that models continuous slice acquisition of the scanner. Under this model the particle filter provides more accurate motion estimates and voxel position estimates. We demonstrate significant performance improvement of the proposed approach as compared to registration-only methods of head motion estimation and brain activation detection.
\end{abstract}

% Note that keywords are not normally used for peerreview papers.

\IEEEpeerreviewmaketitle

\section{Introduction}
Brain activation studies aim to identify specific regions in the brain that are associated with particular tasks. Detection of such functional regions is commonly performed by acquiring functional magnetic resonance imaging (fMRI) data using echo planar imaging (EPI) where the signal contrast is caused by the change of oxygenation in blood flow associated with local upstream neural activity. To detect brain activation in this noisy environment, one typically averages responses over several identical stimuli. Repeated scanning that is synchronous with the onset of the required task (e.g., finger tapping, picture naming, etc.) is used to support signal averaging to improve the signal to noise ratio (SNR) for detecting the blood oxygen level de-saturation (BOLD) response~\cite{turner_functional_1998}. By synchronously averaging the series of brain image volumes over the course of an fMRI study, the BOLD signal contrast can be significantly enhanced.

% In either case the MRI scanner generates a series of images of the subject's head section and the images are concatenated to form the $3$D volume of each scan. The time-series of volumes captures the brain activity signal during the experiment, which serves as the starting point for brain activity analysis.

Ideally, each voxel in the volume time series records the signal evolving over time for a specific position. However, if the head of the subject moves during the scanning process, the time variation of voxel locations results in blurring or loss of signal and severe degradation of the fMRI image. This effect accumulates additional noise in the activation signal, impairing activity analysis accuracy. In experiments that require verbalized activation studies the head cannot be immobilized as the subject is required to speak during scanning. Therefore, some degree of head motion is inevitable even with cooperative subjects.

To deal with the above problem, the head motion can first be estimated and then used to correctly place fMRI image slices into the fMRI volume. Stereo optical tracking systems have been proposed to provide good real-time motion estimation with reasonable accuracy~\cite{zaitsev_magnetic_2006,qin_prospective_2009}. However, these systems require complicated and time-consuming system calibration. Other works use micro radio-frequency coils, called ``active markers'', for real-time prospective correction~\cite{ooi_prospective_2009,ooi_echo-planar_2011}. Although such approaches can achieve good performance, they require additional equipment, incurring additional expense and adding complexity to the experimental protocol. Besides, there is a time lag between the actual instantaneous position of the subject's head and its computation from the sensors. Furthermore, the markers are mounted on the skin whose elasticity can introduce errors in head motion estimation.

\begin{figure}
\begin{center}
\includegraphics[width=0.9\columnwidth]{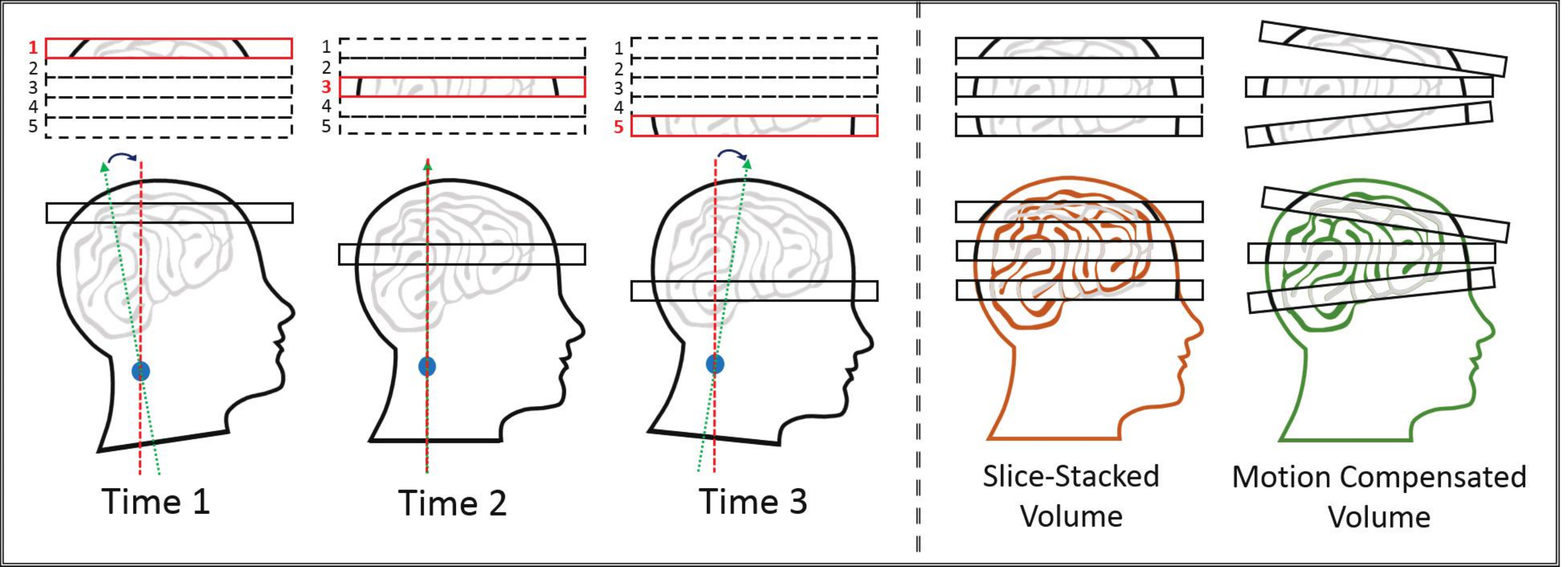}
\caption[]{The inter-slice motion with respect to the scanner caused by head nodding during the scan. Note the interleaved acquisition~\cite{butts_interleaved_1994} shown in the figure that the time interval between adjacent slices is large compared to the slice acquisition interval. The second figure from the right demonstrates the mismatch between the slice-stacked volume and the true human brain due to head motion. The most right figure shows the motion corrected volume, which geometrically instantiates the original brain signal more accurately.}
\label{fig:Notparallel}
\end{center}
\end{figure}

In this paper we take an image registration approach to head motion estimation, which does not require additional equipment. We model the head motion by rigid body motion and the motion is directly estimated from the parameters of a rigid body transformation that maps the target image into a reference image. Specifically, the motion parameters are estimated by optimizing pre-defined image similarity measures, e.g., cross-correlation or mutual information~\cite{maintz_survey_1998}, between functional and anatomical reference images. In~\cite{friston_spatial_1995}, the head motion is estimated for each functional volume by registering the volumes to a reference volume. However, since the EPI images are taken slice by slice, stacking the slices directly and treating them as volumes neglects the head motion between consecutive slices within the same volume, i.e. inter-slice motion. Figure~\ref{fig:Notparallel} shows the inter-slice motion with respect to the scanner caused by head nodding during the scan. Note that, in the interleaved acquisition~\cite{butts_interleaved_1994} as shown in the figure, the time interval between adjacent slices can be multiple of the nominal slice acquisition interval. The second figure from the right demonstrates the mismatch between the slice-stacked volume and the true human brain due to head motion. The most right figure shows how the volume is reconstructed by correcting the motion for each slice, which captures the brain activity signal more accurately. 

Mapping-slice-to-volume (MSV)\cite{kim_motion_1999} proposed by Kim et al. was the first work to address the slice-to-volume registration approach. As compared to volume-to-volume registration, the slice-to-volume approach is capable of estimating and correcting the head motion for each slice by more accurately following the EPI acquisition sequence slice by slice. However, the images at the top apex of the head have fewer image features, and are more prone to geometric distortions~\cite{schmitt_echo-planar_1998} than the slices from the mid brain. This may negatively affect the performance of slice-to-volume registration approaches to motion estimation. A main disadvantage of the slice-to-volume approach is computational complexity as the image similarity measure may have many local maxima in the presence of noise and inadequate image features. As usual, choosing suitable initialization for the optimization process is essential for accurate registration. While the focus of this work is fMRI we wish to acknowledge work focused on fetal anatomical imaging in utero by other authors~\cite{kim_intersection_2010,kainz_fast_2015}.

In this work, we propose a head motion tracking (HMT) algorithm based on a dynamic state space model (SSM) that tracks and estimates the head motion for each slice. The head motion parameters are modeled by a random walk, and the Gaussian particle filter~\cite{kotecha_gaussian_2003} is used to estimate the head motion given the observed sequence of EPI slices. The main advantage of our proposed approach is that it utilizes the information from previous acquired slices to provide a good starting point and effectively reduces the parameter search space in the optimization process, improving registration accuracy. The experimental results in Section~\ref{sec:experiment} show that our approach outperforms other methods in terms of head motion parameter estimation, and in terms of activation detection accuracy for both synthetic and noisy real data.

The paper is organized as follows. In Section~\ref{sec:background}, we review background of the general image registration problem as well as the existing head motion correction methods. In Section~\ref{sec:multimodaltracking}, we describe our Head Motion Tracking (HMT) algorithm and how it is used to estimate the motion parameters. Section~\ref{sec:experiment} shows the experimental results for synthetic and real data, and provides comprehensive comparisons between different approaches. Section~\ref{sec:conclusion} concludes this paper.

\section{Head Motion Estimation by Image Registration}
\label{sec:background}
The aim of image registration is to find a one-to-one transformation $T_\btheta$ that maps a reference image $I_R$ onto a target image $I_T$; The two images which may come from different imaging modalities. The transformation parameter $\btheta$ are found by optimizing an image similarity measure $\bbM(.)$ between the target image and the transformed image $T_\btheta(I_R)$ with respect to $\btheta$:
\begin{equation}
\label{eq:image_registration_objfunc}
\hat{\btheta} = \arg\max_{\btheta}\bbM(I_T, T_\btheta(I_R)),
\end{equation}
where $T_\btheta(.)$ is the transformation function parameterized by $\btheta$. The parameterization of $T_\btheta$ could account for rigid body displacement~\cite{hill_medical_2001}, local deformations~\cite{rueckert_nonrigid_1999}, or other relative differences between the reference and target image volumes~\cite{meyer_demonstration_1997,rohr_landmark-based_2001}. For head motion a rigid body displacement parameterization is adequate: $\btheta = [\alpha, \beta, \gamma, \delta x, \delta y, \delta z]$, where $\alpha,\beta,\gamma$ are spherical Euler angles and $\delta_x,\delta_y,\delta_z$ are spatial positions defining the origin of the spherical coordinate system. 

The image similarity measure used in this paper is the mutual information (MI), which has been widely applied to multi-modal biomedical image registration~\cite{maes_multimodality_1997}. Mutual information between the images can be evaluated by first estimating the marginal and joint distributions $p(X),p(Y),p(X,Y)$ and then substituting into:
\begin{equation}
\label{eq:mutual_information}
I(X;Y)=\sum_{x\in X}\sum_{y\in Y} p(x,y)\log{\frac{p(x,y)}{p(x)p(y)}},
\end{equation}
where $X,Y$ are the random variables of the target and reference images' pixel intensity, respectively.

The image acquisition process starts by collecting an anatomical volume $V_{\mathrm{anat}}$ of the subject's head using $T_1$-weighted MRI~\cite{mcrobbie_mri_2006}, which serves as the reference $I_R$ for a functional MR image. The functional MR images are acquired via multislice single-shot echo-planar imaging (EPI) sequences acquired by $T_2^*$-weighted MRI, which has significantly lower spatial resolution than the $T_1$-weighted MRI. Let $\mathcal{V}=\{V_m\}_{m=1}^M$ denote the set of collected EPI volumes, where $M$ is the total number of volumes acquired during the brain scan session. Each of the EPI volumes is composed of a set of EPI slices $V_m=\{S_{mn}\}_{n=1}^N$, where $N$ is the number of slices per volume. The head motion is estimated by registering the set of EPI images $\mathcal{V}$ onto the anatomical volume $V_{\mathrm{anat}}$. There are two main approaches that are commonly used to perform this multi-modality registration:

\subsubsection{Volume-to-volume Registration} 
Friston et al.~\cite{friston_spatial_1995} proposed to estimate the head motion for each volume by registering the EPI images volume by volume via the optimization:
\begin{equation}
\label{eq:v2v_objFunc}
\hat{\btheta}_m = \arg\max_{\btheta} \bbMI(V_m, T_\btheta(V_{\mathrm{anat}})).
\end{equation}

The advantage of this approach is that the $3$D volume contains abundant image features. However, since the EPI images are acquired slice by slice, this approach is not able to track significant movement occurring between each EPI slice. As EPI slices are commonly acquired in interleaved fashion, the typical time elapsed between adjacent slices can be as large as $1$ second~\cite{turner_functional_1998,butts_interleaved_1994}. Therefore, inter-slice head motion can be significant.

\subsubsection{Slice-to-volume Registration}
Slice-to-volume registration maps each individual slice into the anatomical reference volume space as proposed in~\cite{kim_motion_1999}. The motion parameters are estimated for slices instead of volumes via the optimization:
\begin{equation}
\label{eq:MSV_objFunc}
\hat{\btheta}_{mn} = \arg\max_{\btheta}\bbMI(S_{mn}, T^*_{\btheta}(V_{\mathrm{anat}})),
\end{equation}
where $T^*_{\btheta}(.)$ is the function that interpolates the anatomical volume into $2$D section with the motion parameter $\btheta$. This approach is capable of estimating and recovering the inter-slice head motion. However, because each $2$D EPI slice $S_{mn}$ carries less information than the $3$D volume $V_m$, using (\ref{eq:MSV_objFunc}) can be sensitive to noise. Thus it is important to couple together the registration of successive EPI slices. The coupling of successive EPI slices in the registration process constitutes the main contribution of this paper.

\section{Head Motion Tracking}
\label{sec:multimodaltracking}

\subsection{Coordinate Transformation}
Our head motion tracking algorithm adopts the slice-to-volume approach to estimate the head motion for each EPI slice. As in (\ref{eq:v2v_objFunc}) and (\ref{eq:MSV_objFunc}) we formulate this problem as an optimization. We use a Gaussian particle filter to initialize and track the rigid body motion parameters across EPI slices. Let $\mathcal{S}=\{S_t\}_{t=1}^T$ denote the set of acquired EPI slices re-arranged in order of acquisition time, where $T=MN$ is the total number of slices in the experiment. Given the acquired EPI slices $\mathcal{S}$ and the anatomical volume $V_{\mathrm{anat}}$, the aim of the tracking algorithm is to estimate the head motion parameters at each time $\{\btheta_t\}_{t=1}^T$. Since we model the head motion as a rigid body transformation, the parameter $\btheta_t$ has six degrees of freedom and can be represented as a $3\times 3$ rotation matrix $\bR_t$ and a translation vector $\bq_t$. Let $\bx_r$, $\bx_o$ denote the $3$D-coordinates in the reference and observation coordinate systems. The conversion between the two coordinate systems can be described as:

\begin{equation}
\label{eq:coordinate_conversion}
(\bx_r-\bc) = \bR_t((\bR_s\bx_o+\bq_s)-\bc)+\bq_t,
\end{equation}
where $\bR_s, \bq_s$ are fixed transformations introduced by coordinate mismatch between the two MRI scanners, e.g., due to initial head position difference, and $\bc$ is the head rotation center that ideally corresponds to the location of the cervical vertebrae. Note that $\bR_s, \bq_s, \bc$ are constant over time and only need to be estimated once in the whole experiment. The proposed method to estimate these parameters is discussed in Section~\ref{sec:system_parameters_setting}.

\subsection{Head Motion Tracking Algorithm}
We use a state space model (SSM)~\cite{durbin_time_2012} to describe the head motion, where $\btheta_t$ denotes the rigid body parameters at time $t$. The state equation is modeled using a Gaussian random walk with covariance matrix $\bSigma_d$:

\begin{equation}
\label{eq:state_transition}
\btheta_{t+1} = \btheta_t + \bu_t, ~\bu_t\sim \mathcal{N}(\mathbf 0, \bSigma_d) 
\end{equation}

Note that our HMT algorithm can also be applied with more general head motion model, e.g., a kinematic model~\cite{han_visual_2009}. The acquired EPI slices, called the observations in the sequel, are related to the state through the quasi-likelihood function:

\begin{equation}
\label{eq:likelihood}
p(\bS_t|\btheta_t) = \frac{1}{Z}\bbL(\bbM(\bS_t,T_{\btheta_t}^*(V_{\mathrm{anat}}))),
\end{equation}
where $\bbL(.)$ can be chosen as any function such that it is positive and monotonically increasing (i.e. $\bbL(x)\ge 0,~\forall -\infty<x<\infty, x>y\Rightarrow \bbL(x)>\bbL(y)$) and $Z$ is a normalization coefficient that turns the objective function $\bbL(.)$ into a conditional probability, which is denoted $p(\bS_t|\btheta_t)$ and is called the quasi-likelihood function of $\btheta_t$. Here $\bS_t=\{S_j\}_{j=t-h}^{t+h}$ denotes the stack of slices over a length $2h+1$ time interval centered at time $t$. If $h=0$, $\bS_t$ is reduced to a single EPI slice $S_t$. The parameter $h$ controls the trade-off between parameter estimator bias and variance. In the analysis reported below we have used $h=1$, which was found to achieve a good trade-off between these two factors.

The Kalman Filter~\cite{kalman_new_1960} is the optimal minimum mean squared error estimator for a linear SSM when both the state dynamics and the measurement equations are linear in the state vector and the driving noise vectors. In non-linear cases, one has to resort to some form of approximation to the minimum mean squared error estimator, such as the extended Kalman filter (EKF)\cite{julier_new_1997} or the unscented Kalman filter (UKF)\cite{wan_unscented_2000}. These approaches require explicit state and observation equations, which are not readily available in the fMRI problem treated here. Alternatively, one can approximate the posterior distribution of the state using sequential importance sampling, i.e., the particle filter~\cite{doucet_sequential_2000}.

The Gaussian particle filter (GPF)\cite{kotecha_gaussian_2003} approximates the posterior using a set of weighted samples, known as particles, and uses importance sampling and Monte-Carlo integration methods to approximate the state and observation distributions. The main advantage of GPF compared to other particle filtering approaches is its lower computational complexity and amenability to parallel implementation. In GPF algorithm, the posterior at time $t$ is approximated by a Gaussian distribution $\mathcal{N}(\bmu_t, \bSigma_t)$, and then resampling follows by drawing $P$ particles from the Gaussian distribution. The particles are weighted according to the observation and are used to form the distribution for the next time step. 

Our Head Motion Tracking (HMT) algorithm is based on the GPF framework which is summarized in Algorithm~\ref{algo:HMT_algo}. Initially slice-to-volume registration~\cite{kim_motion_1999} is used to generate an initial head motion estimate $\hat{\btheta}_0$. As in the GPF, for each slice at time $t$, the algorithm has two stages: \emph{Measurement update} and \emph{Time update}. In the \emph{Measurement update} stage, we use $P$ particles $\{\btheta_t^{(j)}\}_{j=1}^P$ drawn at the last time step to evaluate the particle weights using the quasi-likelihood function $p(\bS_t|\btheta_t)$ defined in (\ref{eq:likelihood}). The quasi-likelihood function should have two properties: (1) It is monotonically increasing with the image similarity $\bbM(\bS_t, T_\btheta^*(V_{\mathrm{anat}}))$; (2) The weighted particles are approximately distributed according to a multivariate Gaussian density. To satisfy the two properties, we propose to use a histogram equalization approach to evaluate the particle weights. The target density is the distribution of $z=f(\bx)$ where $\bx$ and $f(.)$ are the $6$-dimension multivariate Gaussian random variable and density, respectively. Letting $g_Z(z)$ denote the density of $z$, we can equalize the histogram to obtain the particle weights. 
\begin{equation}
\label{eq:gz_d6_direct}
g_Z(z)=\pi^3\left(-2\log{(2\pi)^3z}\right)^2,~z\in (0,(2\pi)^{-3}].
\end{equation}

The detailed derivation of~(\ref{eq:gz_d6_direct}) is given in Appendix~\ref{append:particle_weight_evaluation}. The particle weights are normalized to sum to $1$ and then used to calculate the weighted mean and covariance. Since the weighted mean incorporates abundant information about the image similarity distribution in neighboring regions, it is a good starting point for the optimizer. In this paper, we use the Nelder-Mead~\cite{nelder_simplex_1965} optimizer to perform the maximization:
\begin{equation}
\label{eq:HMT_obj_func}
\hat{\btheta}_t=\arg\max_{\btheta}~\bbM(\bS_t, T_{\btheta}^*(V_{\mathrm{anat}})).
\end{equation}

Nelder-Mead is a simplex method used to iteratively find the optimum of an objective function in a multi-dimensional space. Note that the proposed histogram equalization approach is not restricted to any particular definition of image similarity. Therefore MI can be replaced by any other image similarity measure, e.g., Normalized MI~\cite{studholme_overlap_1999}, localized MI~\cite{klein_automatic_2008}, graph-based MI~\cite{staring_registration_2009}, or feature-based measures~\cite{oliveira_medical_2014}...etc. The transformation parameter $\hat{\btheta}_t$ that maximizes (\ref{eq:HMT_obj_func}) is the estimated head motion at time $t$. After the motion parameter is estimated, we perform a standard re-sampling step to estimate the covariance matrix of the posterior distribution, which is then used to establish the prior distribution of the next slice in the \emph{Time Update} stage using (\ref{eq:state_transition}).

\begin{figure}[ht]
\begin{center}
\subfloat[middle slice of head]{{\includegraphics[width=0.5\columnwidth]{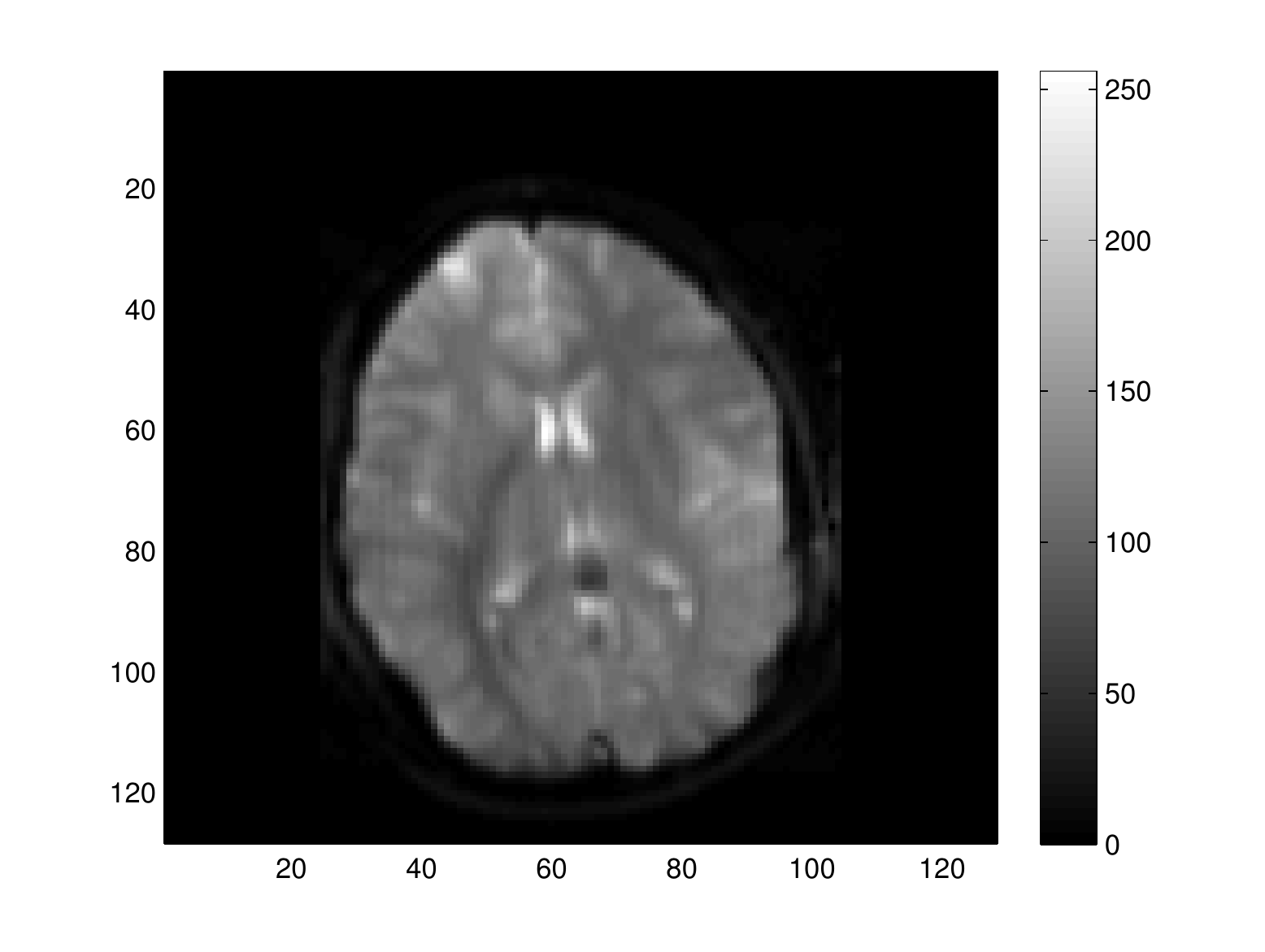}}}
\subfloat[top apex slice of head]{{\includegraphics[width=0.5\columnwidth]{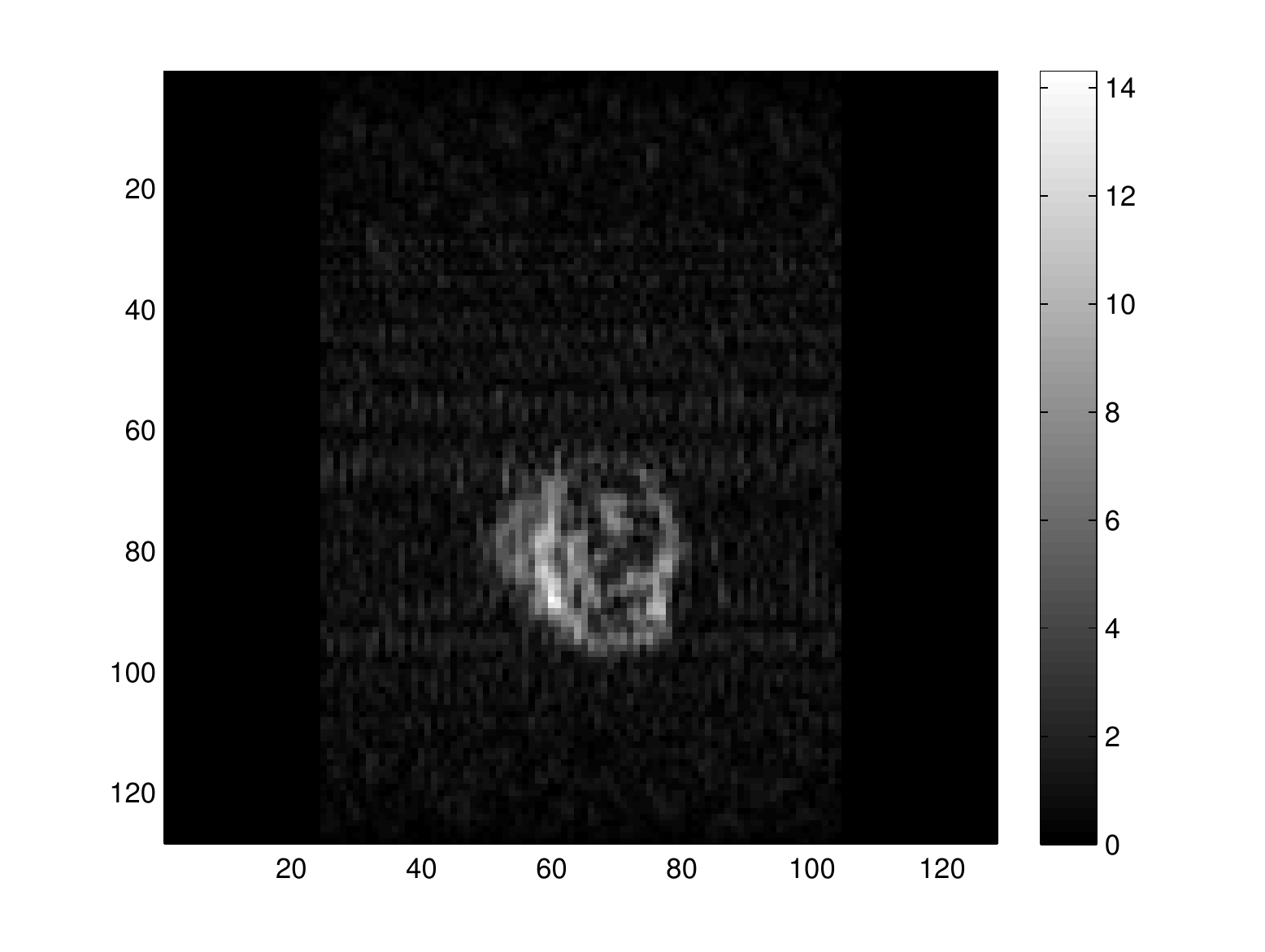}}}
\end{center}
\caption{The middle head (a) and top apex (b) of the real human data are shown in gray scale. Notice that the top apex image has very little useful features and the signal strength (pixel intensity) is much lower than the middle head image.}
\label{fig:real_slc_examples}
\end{figure}

Often the MRI acquired images are very noisy and difficult to register, especially for slices near the lower and upper apex of the head. Figure~\ref{fig:real_slc_examples} shows an example of the images of the middle head (a) and top apex (b). We can see that the top apex head image has much less information content than the middle head that can be used for registration. To reduce the effect of these noisy slices, we screen the slices for adequate signal strength. Specifically, we reject all EPI slices for which fewer than $15\%$ of the pixels are above a certain threshold value. For these rejected slices, we skip the optimization step and estimate the motion parameters through interpolation of the estimates from neighboring slices. We use $2$nd-order interpolation, which is accurate when the head motion has approximately constant angular and translational accelerations~\cite{park_improved_2004}.

\subsection{System Parameters Setting}
\label{sec:system_parameters_setting}
In the proposed Head Motion Tracking algorithm there are several parameters that need to be set: $\bR_s, \bq_s, \bc, \bSigma_d$:

\subsubsection{Fixed Coordinate Transformation $\bR_s,\bq_s$}
Since $\bR_s,\bq_s$ are constant over the entire experiment, they can be estimated by first taking the average of all EPI volumes over time, and then registering the averaged EPI volume to the anatomical volume.

\subsubsection{Head Rotation Center $\bc$}
To estimate the head rotation center, we run the HMT algorithm on the first $K$ image slices (we used $K=70$ in our experiment) by assuming $\bc=\mathbf{0}$ as the origin. Let $\{\hat{\btheta}_t\}_{t=1}^K$ denote the estimate of the motion parameters for these $K$ image slices. Here we assume that the patient's body position is stable during the scan (the subject is often immobilized and lying in the machine) and therefore the amount of translation should be small, i.e. $\|\bq_t\|\approx 0$. Based on this assumption, the rotation center can be estimated by solving the least squares problem:
\begin{equation}
\label{eq:c_estimation}
\hat{\bc} = \arg\min_\bc \sum_{t=1}^K\|\bq_t-(\bI_3-\bR_t)\bc\|^2_2,
\end{equation}
where $\bI_3$ is the $3\times3$ identity matrix.

\subsubsection{Head Motion Covariance $\bSigma_d$}
The estimate of the head motion covariance matrix is generated in two steps. We initially set $\bSigma_d$ to the identity matrix and run the HMT algorithm over $K$ image slices to obtain the estimates $\{\hat{\btheta}_t\}_{t=1}^K$. Subsequently, the matrix $\bSigma_d$ is estimated as the covariance matrix of the consecutive parameter differences:
\begin{equation}
\label{eq:sigma_d_estimation}
\begin{split}
\hat{\bSigma}_d &= Cov(\hat{\btheta}_t-\hat{\btheta}_{t-1}) \\
&= \frac{1}{K-1}\sum_{t=2}^K (\hat{\btheta}_t-\hat{\btheta}_{t-1})(\hat{\btheta}_t-\hat{\btheta}_{t-1})^T
\end{split}
\end{equation}

\begin{algorithm}{HMT}{
\label{algo:HMT_algo}
\qinput{EPI slices $\{S_t\}_{t=1}^{T}$ and anatomical volume $V_{\mathrm{anat}}$}
}
Estimate the parameters for the first slice $\hat{\btheta}_0$ using slice-to-volume registration. \\
Draw $P$ particles $\{\btheta_0^{(j)}\}_{j=1}^P$ from $\mathcal{N}(\hat{\btheta}_0, \bSigma_d)$. \\
\qfor $t \qlet 1$ \qto $T$ \\
\qcom{\emph{Measurement update}} \\ 
Equalize the histogram of $\bbM(\bS_t, T_{\btheta_t^{(j)}}^*(V_{\mathrm{anat}}))$ to (\ref{eq:gz_d6_direct}) to get $\bar{w}_t^{(j)}$ and then normalize to sum to $1$

\begin{equation}
w_t^{(j)}=\bar{w}_t^{(j)} / \sum_{j=1}^P\bar{w}_t^{(j)} \nonumber
\end{equation} \\
Estimate the sample mean and covariance 
\begin{equation}
\label{eq:weighted_mu_sigma}
\begin{split}
\bmu_t &= \sum_{j=1}^P w_t^{(j)}\btheta_t^{(j)} \\
\bSigma_t&=\sum_{j=1}^Pw_t^{(j)}(\btheta_t^{(j)}-\bmu_t)(\btheta_t^{(j)}-\bmu_t)^T \\
\end{split} \nonumber
\end{equation} \\
Initialize the registration process with $\bmu_t$ to estimate the motion parameter:
\begin{equation}
\hat{\btheta}_t=\arg\max_{\btheta}\bbM(\bS_t, T_{\btheta}^*(V_{\mathrm{anat}})) \nonumber
\end{equation} \\

\qcom{\emph{Time update}} \\
Draw samples $\{\btheta_t^{(j)}\}_{j=1}^P$ from $\mathcal{N}(\btheta_t, \bSigma_t)$. \\
For $j=1,...,P$, sample from $p(\btheta_{t+1}|\btheta_t=\btheta_t^{(j)})$ to obtain $\{\btheta_{t+1}^{(j)}\}_{j=1}^P$.\\

\qreturn $\{\hat{\btheta}_t\}_{t=1}^T$
\end{algorithm}

\section{Experimental Results}
\label{sec:experiment}
\subsection{Synthetic Data Generation}
\label{sec:experiment_synthetic}
We downloaded high resolution $T_1$, $T_2$-weighted MRI volumes from the International Consortium of Brain Mapping (ICBM)~\cite{cocosco_brainweb:_1997}. The high resolution $T_1$ MRI brain volume was used as the anatomical reference volume with voxel size $0.78\times 0.78\times 1.5mm^3$. The EPI slices were emulated by interpolating the $T_2$-weighted volume under artificial motion induced by applying a sequence of transformations to the image with smoothly varying motion parameters. The voxel size of the EPI slices is $1.56\times 1.56\times 6mm^3$, a blurring Gaussian low-pass kernel with $\sigma=2$ was applied, and $3\%$ Gaussian noise was added to simulate real EPI slices. The activation signal was introduced by adding $5\%$ intensity to selected voxels in manually drawn regions of interest at various locations in the volume as in~\cite{kim_comprehensive_2008}. This produced a synthetic EPI data set consisting of $M=120$ volumes with $N=14$ slices per volume. Figure~\ref{fig:estimated_pars_simulation}(a) shows the ground truth motion parameter of the three rotation angles (in degree) from slice $1$ to $200$. The simulated time series in a block design paradigm consists of $120$ volumes with $6$ activation cycles. There are $20$ volumes per cycle which contains $10$ stimulation and $10$ control volumes.

\subsection{Performance Measures}
In the following comparison, we evaluate the performance quantitatively with respect to misregistration error, activation detection accuracy and reliability:

\subsubsection{Average Voxel Distance}
The misregistration error is measured by \emph{average voxel distance}, which is the average distance between the registered voxel coordinate and the true voxel coordinate. Let $\bx_t^{reg}(i)$ and $\bx_t^{true}(i)$ denote the coordinates of voxel $i$ transformed using the estimated motion parameter $\hat{\btheta}_t$ and true motion parameter $\btheta_t$ of slice $t$. The average voxel distance is defined as:

\begin{equation}
\label{eq:avg_voxel_distance}
D_t = \frac{1}{N_v}\sum_{i=1}^{N_v} \|\bx_t^{reg}(i)-\bx_t^{true}(i)\|,
\end{equation}
where $N_v$ is the total number of voxels in a single EPI slice.

\subsubsection{Activation Detection ROC Curve}
\label{sec:Act_ROC}
The estimated motion parameters $\{\hat{\btheta}_t\}_{t=1}^T$ are used to reconstruct the motion corrected EPI volumes $\tilde{\mathcal{V}}=\{\tilde{V}_m\}_{m=1}^M$. To identify the activated brain region, the non-parametric random permutation test\cite{nichols_nonparametric_2002} is performed on the intensities in the EPI volumes. Let $\{u_m(i)\}_{m=1}^M$ to be the set of intensities for voxel $i$ of the $M$ reconstructed volumes. The null hypothesis $H_0$ of the activation test is: "The mean of the voxel intensities under each of the conditions, stimulation or control, are equal." Under this hypothesis, any re-ordering of $\{u_m(i)\}_{m=1}^M$ should give the same statistic, which we used the two-sample $t$-test statistic. Let $N_r$ denote the number of re-ordering, $t_j$ be the two-sample $t$-test statistic corresponding to ordering $j$ and $\tilde{t}$ be the statistic of actual ordering. The $P$-value is then calculated by counting the proportion of the test statistics $\{t_j\}_{j=1}^{N_r}$ which are more extreme than $\tilde{t}$. By taking a threshold on the $P$-value, we can determine which voxels are activated in this experiment. In this paper, we set $N_r$ equal to $2000$ and the threshold for $P$-value is $0.5\%$. When a ground truth activation map is available as in the synthetic data, the detection performance can be evaluated by the Receiver Operating Characteristic (ROC) curve and the Area Under the Curve (AUC). 

\subsubsection{Activation Detection Reliability}
We use the Activation Test-retest Reliability (ATR) measure~\cite{noll_estimating_1997,genovese_estimating_1997} to compare the performance when the ground truth of motion parameters and activation map are unknown. This approach assumes that each voxel is either truly active or truly inactive. We use the random permutation test with two-sample $t$-test statistic to generate the activation maps as describe in Section~\ref{sec:Act_ROC}. The reliability of the test is measured in terms of true active and false active probability, $p_A=p(\{v\text{ is classified as active}\}|\allowbreak\{v\text{ is truly active}\})$ and $p_I=p(\{v\text{ is classified as active}\}|\allowbreak\{v\text{ is truly inactive}\})$, respectively. Ideally, $p_A$ should be $1$ and $p_I$ should be $0$. Therefore, higher $p_A$ and lower $p_I$ indicate more reliable testing result. 

To estimate $p_A,p_I$, we need to replicate the fMRI experiments $L$ times, where $L\ge3$. In this paper, we obtain the replications by splitting the acquired volumes into $L=4$ disjoint sets randomly as suggested in~\cite{noll_estimating_1997} to ensure statistical independence accross voxels and replications. We use the random permutation test to generate $L$ activation maps for each of the sets. Let $r(i)\in\{0,1,...,L\}$ be the number of replications out of $L$ in which the voxel $i$ is classified active. We model $r(i)$ as a mixture of two binomial distributions:
\begin{equation}
\label{eq:atr_bino_model}
\lambda \mathbf{B}(L,p_A)+(1-\lambda)\mathbf{B}(L,p_I),
\end{equation}
where $\mathbf{B}$ is the binomial distribution and $\lambda$ represents the proportion of truly active voxels. We estimated the parameters by maximizing the likelihood function.

\subsection{Evaluation Using Synthetic Data}
\label{sec:Eval_Syn}
\begin{figure}[ht]
\begin{center}$
\begin{array}{cc}
\subfloat[Ground Truth]{{\includegraphics[width=0.47\columnwidth]{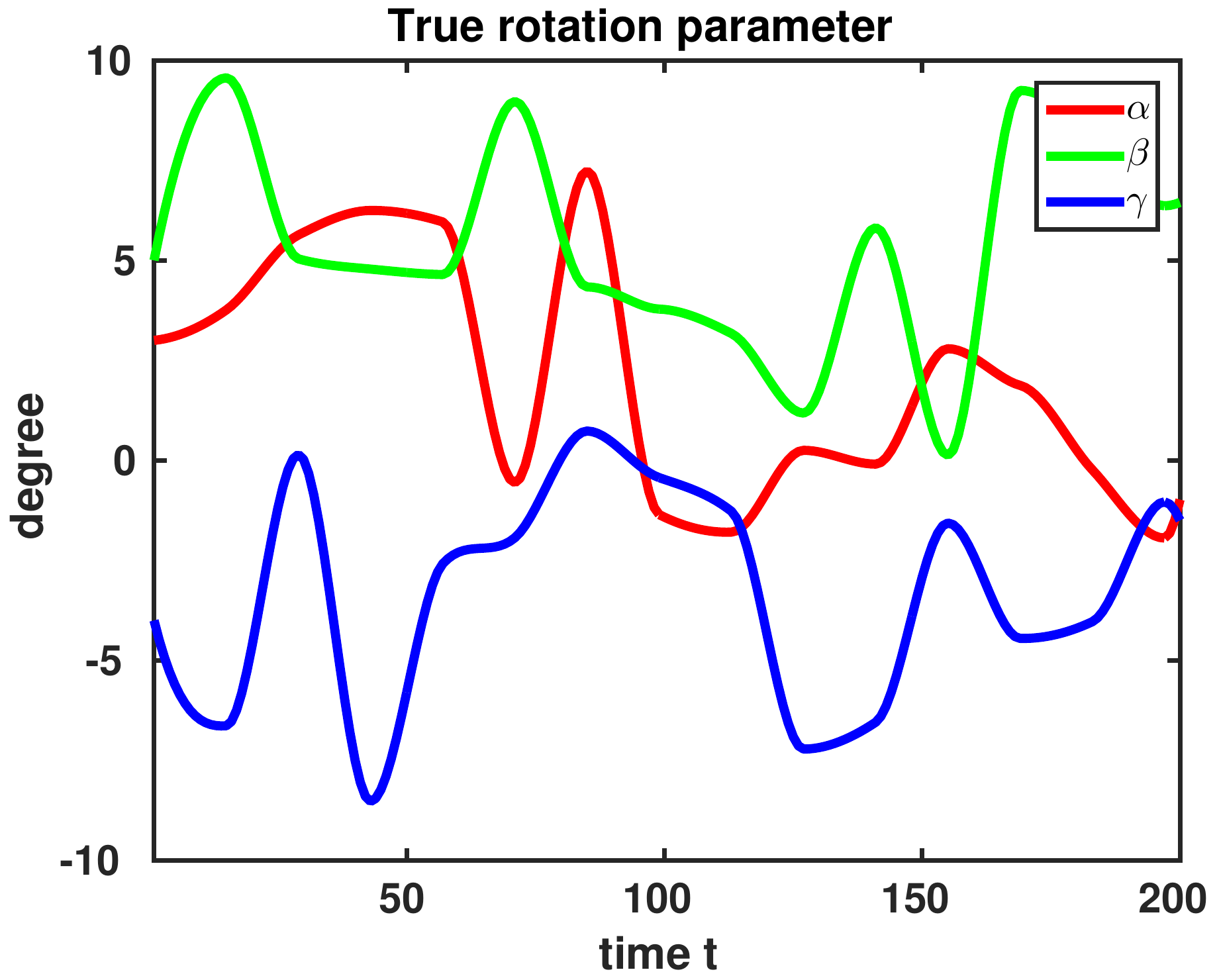}}} &
\subfloat[V2V]{{\includegraphics[width=0.47\columnwidth]{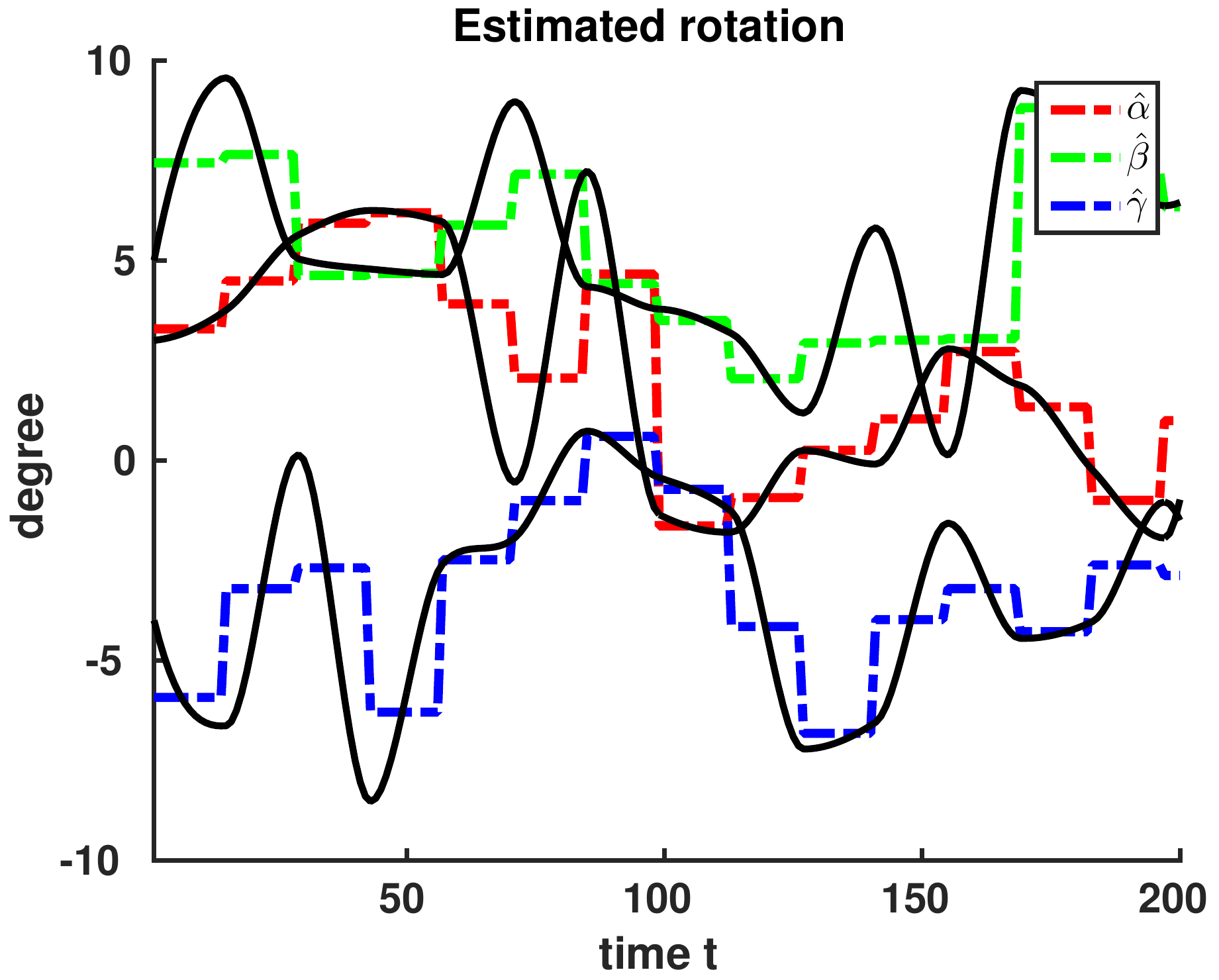}}} \\
\subfloat[S2V]{{\includegraphics[width=0.47\columnwidth]{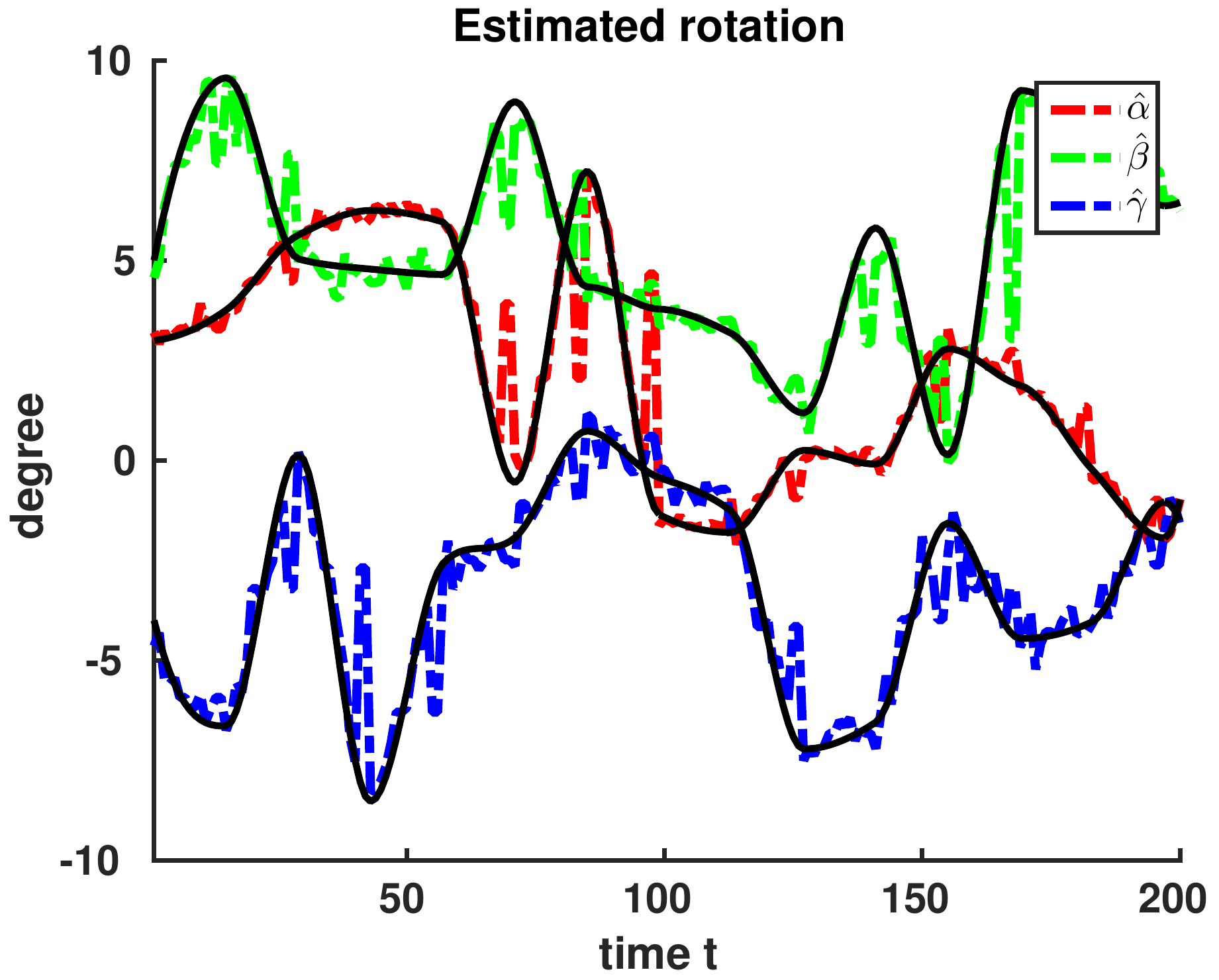}}} &
\subfloat[HMT]{{\includegraphics[width=0.47\columnwidth]{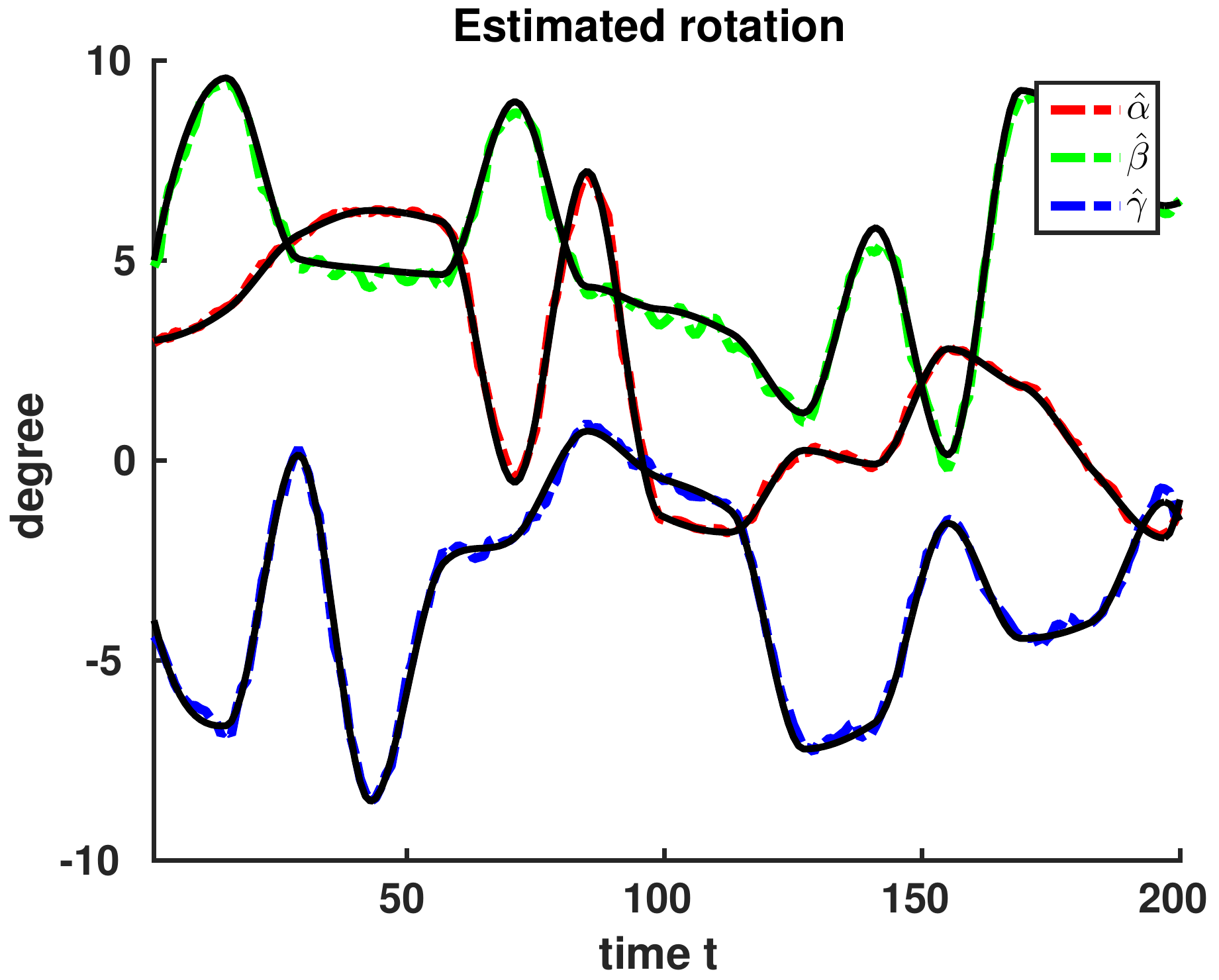}}}
\end{array}$
\end{center}
\caption{(a) shows the ground truth of head motion parameters in three Euler angles for the first $200$ slices. (b)(c)(d) show the motion parameters estimated by volume-to-volume (V2V), slice-to-volume (S2V) and the proposed head motion tracking (HMT) algorithm. The black solid lines are the ground truth and the color dashed lines are the estimated motion parameters. (b) demonstrates that the volume-to-volume registration method can accurately track the average motion for each volume but does not accurately track motion for each slice in the volume. S2V (c) can estimate the head motion for each slice but suffers from large tracking errors. The proposed HMT algorithm (d) is able to track the head motion much accurately than the other two approaches.}
\label{fig:estimated_pars_simulation}
\end{figure}

\begin{figure}[ht]
\begin{center}$
\begin{array}{cc}
\subfloat[Avg. Voxel Distance]{{\includegraphics[width=0.47\columnwidth]{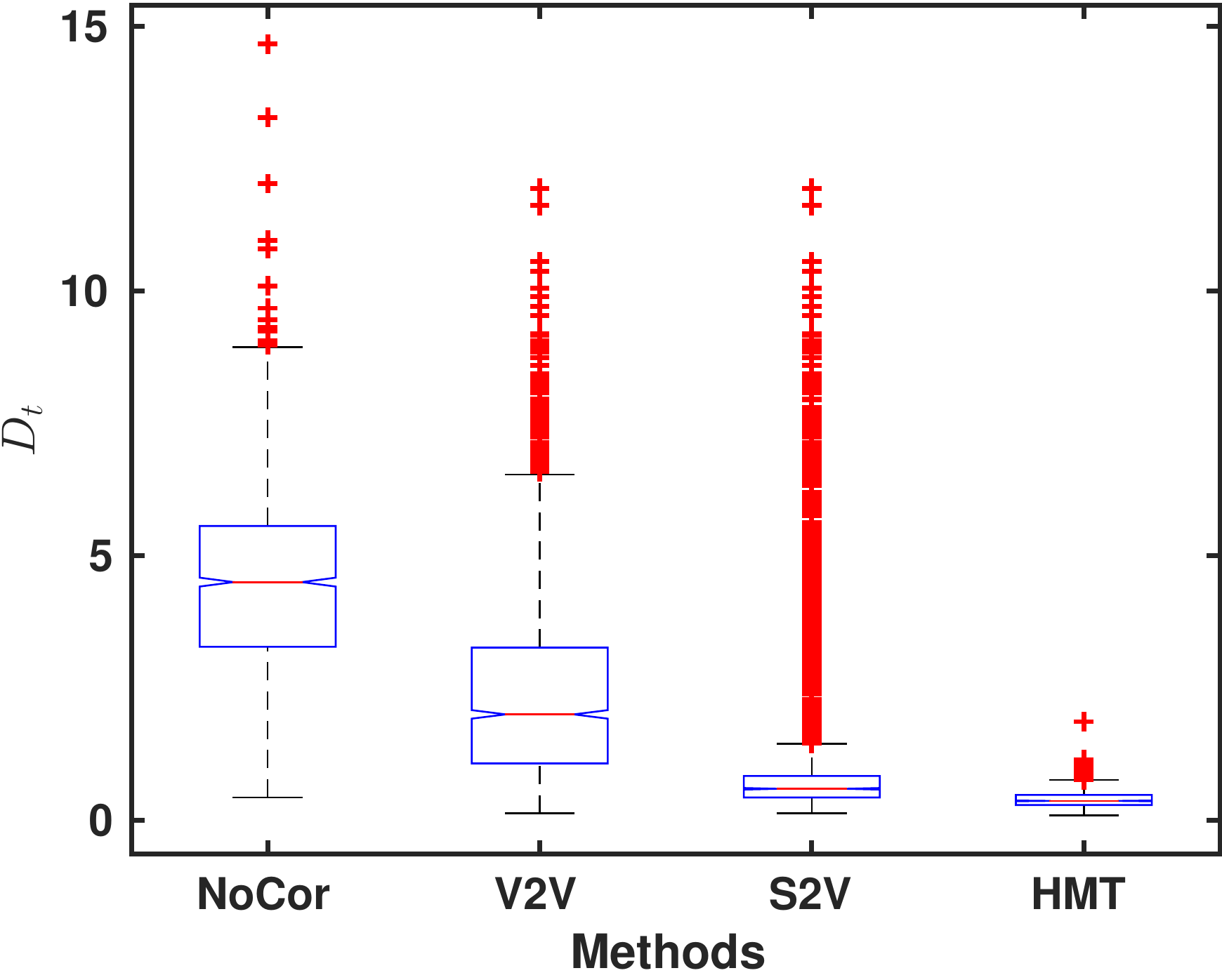}}} & 
\subfloat[Activation ROC Curve]{{\includegraphics[width=0.47\columnwidth]{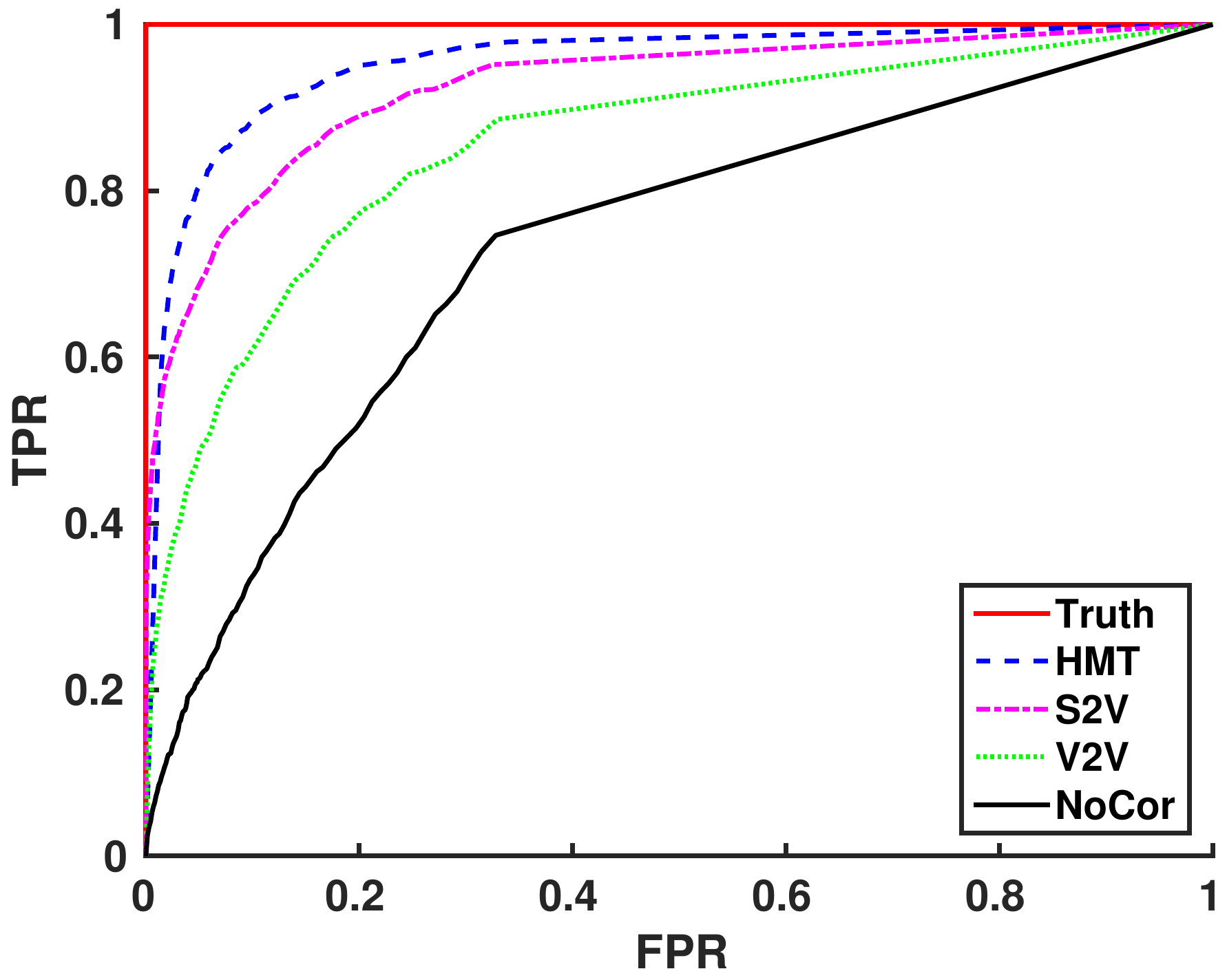}}} 
\end{array}$
\end{center}
\caption{(a) is the boxplot of the average voxel distance after registration for different methods. The whiskers are the outliers outside the inner fence (defined by $1.5\times F$-spread). The proposed HMT algorithm has significantly lower voxel misregistration errors and is more stable (fewer outliers) than the other methods. (b) shows the ROC curves for activation detection. Note that the volumes that are reconstructed using ground truth motion parameters achieve perfect detection (red solid line). Our proposed HMT algorithm (blue dashed line) outperforms other methods (S2V, V2V, No Correction) and is closest to the ground truth.}
\label{fig:avg_disp_roc}
\end{figure}

The simulated EPI slices described in Section~\ref{sec:experiment_synthetic} are registered to the anatomical volume to estimate the motion parameters by using the following three methods (implemented in MATLAB R2015a): (1) volume-to-volume registration~\cite{friston_spatial_1995} (V2V); (2) slice-to-volume registration~\cite{kim_motion_1999} (S2V), where the optimization process is initialized by the V2V result; (3) the proposed Head Motion Tracking algorithm (HMT) with $P=4000$ particles. Figures~\ref{fig:estimated_pars_simulation}(b)-(d) show the estimated motion parameters for the first $200$ slices, where the black solid lines denote ground truth and the color dashed lines denote estimated motion parameters. Figure~\ref{fig:estimated_pars_simulation}(b) demonstrates that the volume-to-volume registration method can accurately estimate motion for each volume but cannot accurately track the motion over the slices. On the other hand, S2V (Fig.~\ref{fig:estimated_pars_simulation}(c)) can better track the head motion over different slices but has high bias, especially for slices near the apex of the head where slice image intensity and contrast are low. Our proposed HMT algorithm (Fig.~\ref{fig:estimated_pars_simulation}(d)) is able to track the head motion much more accurately than the other two approaches. Figure~\ref{fig:avg_disp_roc}(a) shows the boxplot of the average voxel distance after registration for different methods. The whiskers are the outliers outside the inner fence (defined by $1.5\times F$-spread~\cite{hoaglin_performance_1986}). All of these methods reduced a fair amount of the voxel misregistration errors compared to no motion correction case (NoCorr). Notice that our HMT algorithm has significantly lower misregistration error, as measured by voxel distance, and is much more stable (fewer outliers) than the other methods. The mean of $D_t$ over all slices are listed in the first column of Table~\ref{table:Act_Comp}.

The estimated parameters are used to reconstruct the motion corrected EPI volumes, and activated voxels are identified by the random permutation test. The ROC curves of the activation detection result of different approaches are compared in Fig.~\ref{fig:avg_disp_roc}(b). Note that the volumes that are reconstructed using ground truth motion parameters achieve perfect detection (red solid line). Again, our HMT algorithm (blue dashed line) outperforms other methods and is closest to the ground truth. The Area under Curve (AUC) for each approach is listed in the second column of Table~\ref{table:Act_Comp}. The comparison of activation detection reliability is listed in the first two columns in Table~\ref{table:ATR}. It can be seen that all of the three methods have similar $p_I$, but the proposed HMT has significantly higher $p_A$ than the other two methods.

\begin{table}
\begin{center}
\caption{Estimation and Activation Result Comparison}
\label{table:Act_Comp}
\begin{tabular}{|l|l|l|}
\hline%\noalign{\smallskip}
	& Avg. $D_t$ & AUC  \\
%\noalign{\smallskip}
\hline
\hline
%\noalign{\smallskip}
Truth   	& 0.000	& 1.000  \\
No Corr. 	& 4.497	& 0.732  \\
V2V     	& 2.426	& 0.855  \\
S2V     	& 1.225	& 0.924  \\
HMT      	& \textbf{0.393}	& \textbf{0.953} \\
\hline
\end{tabular} 
\\
\smallskip
\justify
As compared to the other motion compensation algorithms (No Corr., V2V, S2V), the proposed HMT algorithm attains lower average misregistration error $D_t$ and better Area Under the Curve (AUC) detection performance.
\end{center}
\end{table}

\subsection{Evaluation Using Real Data}
We further validate the performance of the proposed HMT algorithm on real fMRI experimental data. We used two datasets that are denoted "Run1" and "Run2", and that were acquired from two normal volunteers. The study was approved by the Institutional Review Board at the University of Michigan Medical School and informed consent was obtained from each subject prior to participation. The subjects performed a simple motor task, uni-lateral sequential finger tapping, in the experiment. We asked the subject to do their best to minimize head motion for Run1 dataset and asked the subject to intentionally nod his head for Run2 dataset. The head was scanned $126$ times with $14$ slices in each volume for these two datasets. The anatomical voxel size is $1\times 1\times 1.5mm^3$ and the EPI voxel size is $2\times 2\times 6mm^3$.

\begin{figure}[ht]
\begin{center}$
\begin{array}{cc}
\subfloat[Run1: S2V]{{\includegraphics[width=0.46\columnwidth]{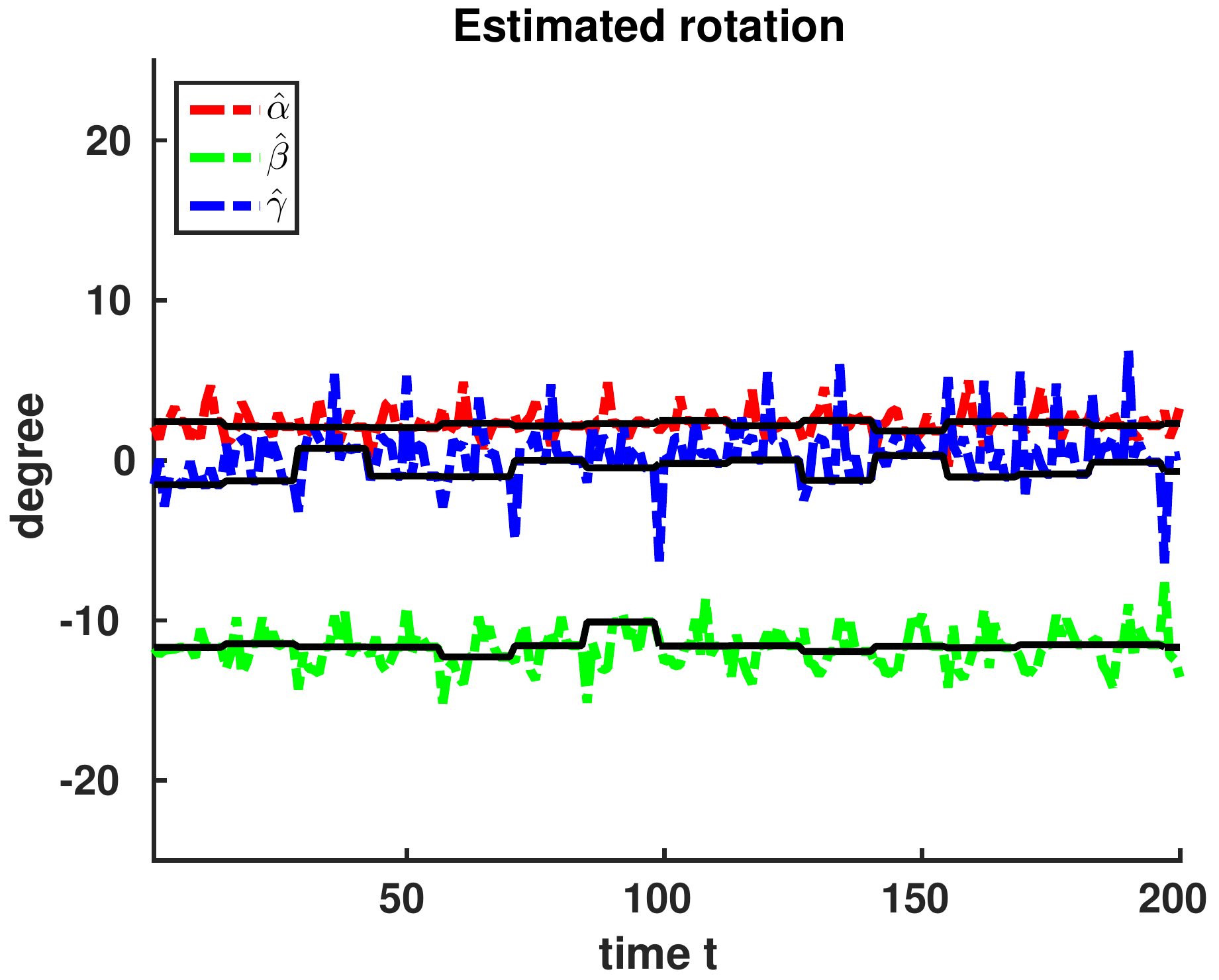}}}&
\subfloat[Run1: HMT]{{\includegraphics[width=0.46\columnwidth]{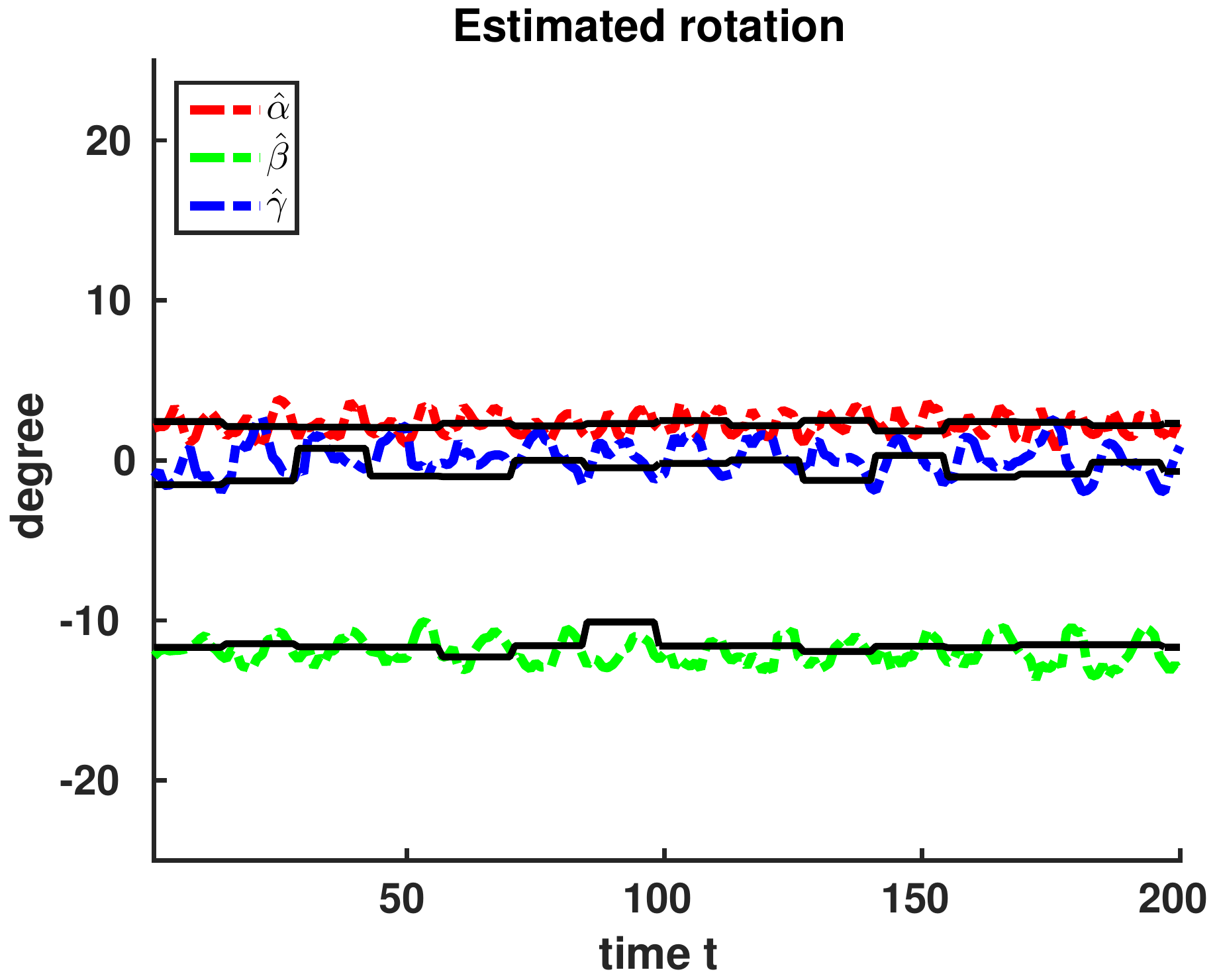}}} \\
\subfloat[Run2: S2V]{{\includegraphics[width=0.46\columnwidth]{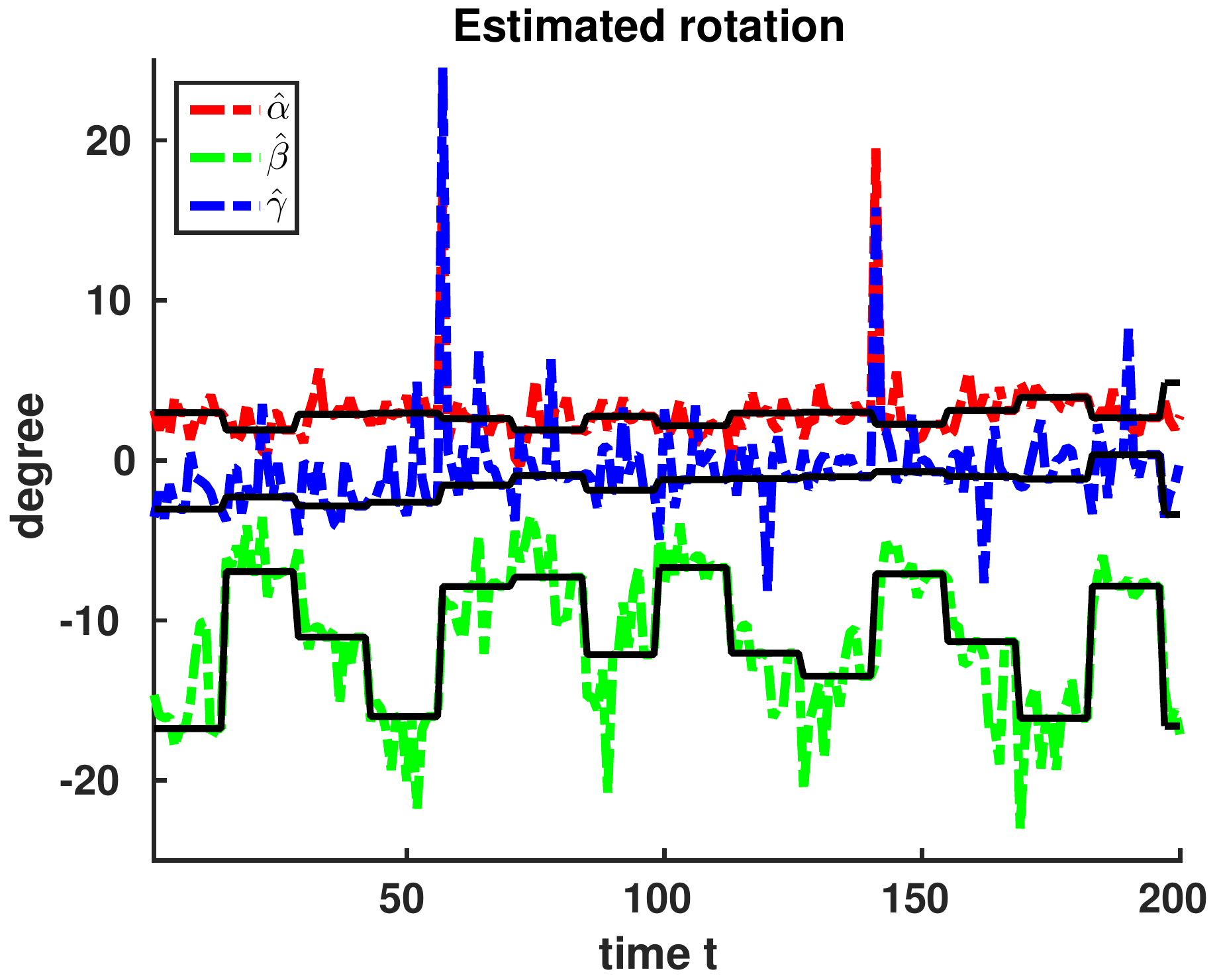}}}&
\subfloat[Run2: HMT]{{\includegraphics[width=0.46\columnwidth]{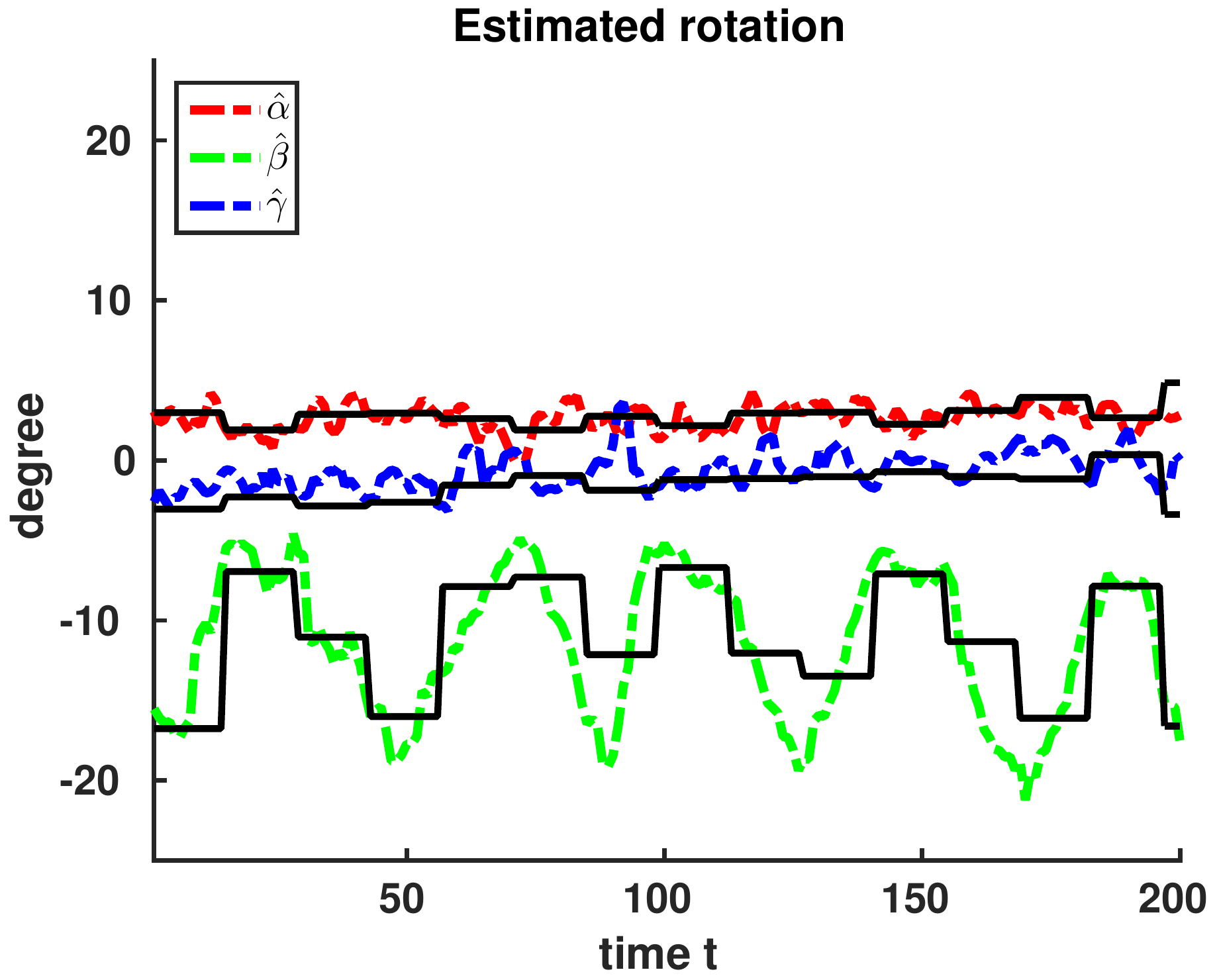}}} 
\end{array}$
\end{center}
\caption{The three Euler angles estimated by S2V (first column color dashed lines) and HMT (second column color dashed lines) overlaid with V2V result (black solid lines) for the first $200$ slices. Note that the estimated rotation in Run2 (second row) is larger than that of Run1 (first row). Similarly to the experiments with synthetic data summarized in Fig.~\ref{fig:estimated_pars_simulation}, S2V can estimate the motion for each slice but is noisy. The proposed HMT algorithm produces more stable and continuous head motion estimates which is more convincing in describing real head motion.}
\label{fig:Real_Est}
\end{figure}

Figure~\ref{fig:Real_Est} shows the three Euler angles estimated by S2V (first column color dashed lines) and HMT (second column color dashed lines) overlaid with the V2V result (black solid lines) for the first $200$ slices. Notice that the estimated rotation in Run2 (second row) is larger than Run1 (first row), which matches our expectations given the experimental protocol. Similarly to the experiments with synthetic data, reported in Section~\ref{sec:Eval_Syn}, S2V can be used to estimate the motion for each slice but is noisy. The abrupt changes in the motion parameters demonstrated by S2V represent unlikely head movement, which suggests incorrect estimation. On the other hand, the proposed HMT algorithm produced much more stable and smoother motion estimates, which more accurately reflects real head motion. The superior tracking performance of HMT is a consequence of the dynamical modeling that couples together estimates from successive slices leading to smoother and less noisy tracking performance.

\begin{figure*}[ht]
\hspace{-0.02\textwidth}
\begin{minipage}[c]{0.49\textwidth}
\begin{center}
\subfloat[Run1 Activation Map]{{
\setlength\arraycolsep{\imgspacing}
$\begin{array}{cccc}
\rotatebox{90}{\hspace{20pt}Slice A}
&\includegraphics[width=\imgwidth, height=\imgheight]{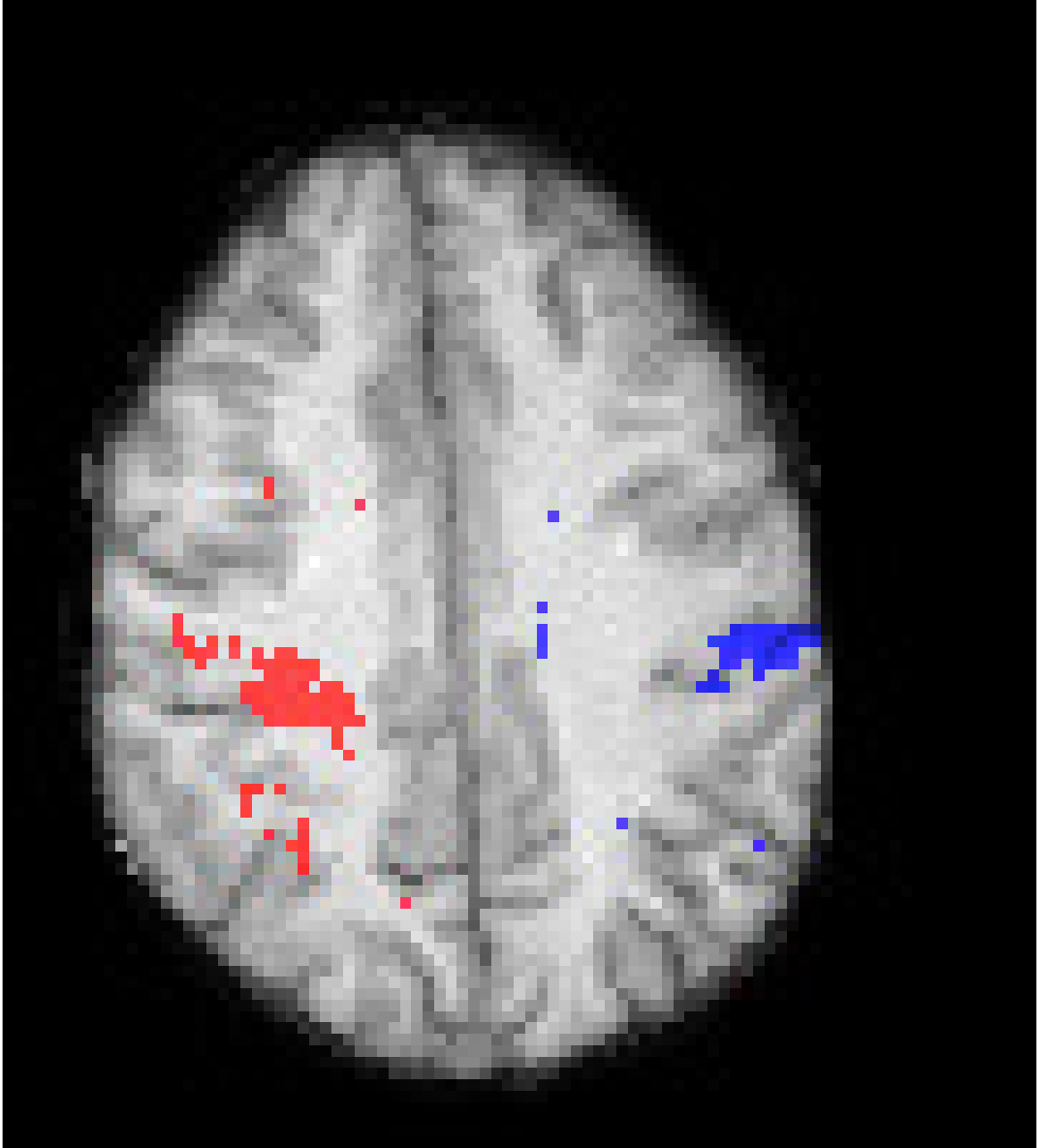}
&\includegraphics[width=\imgwidth, height=\imgheight]{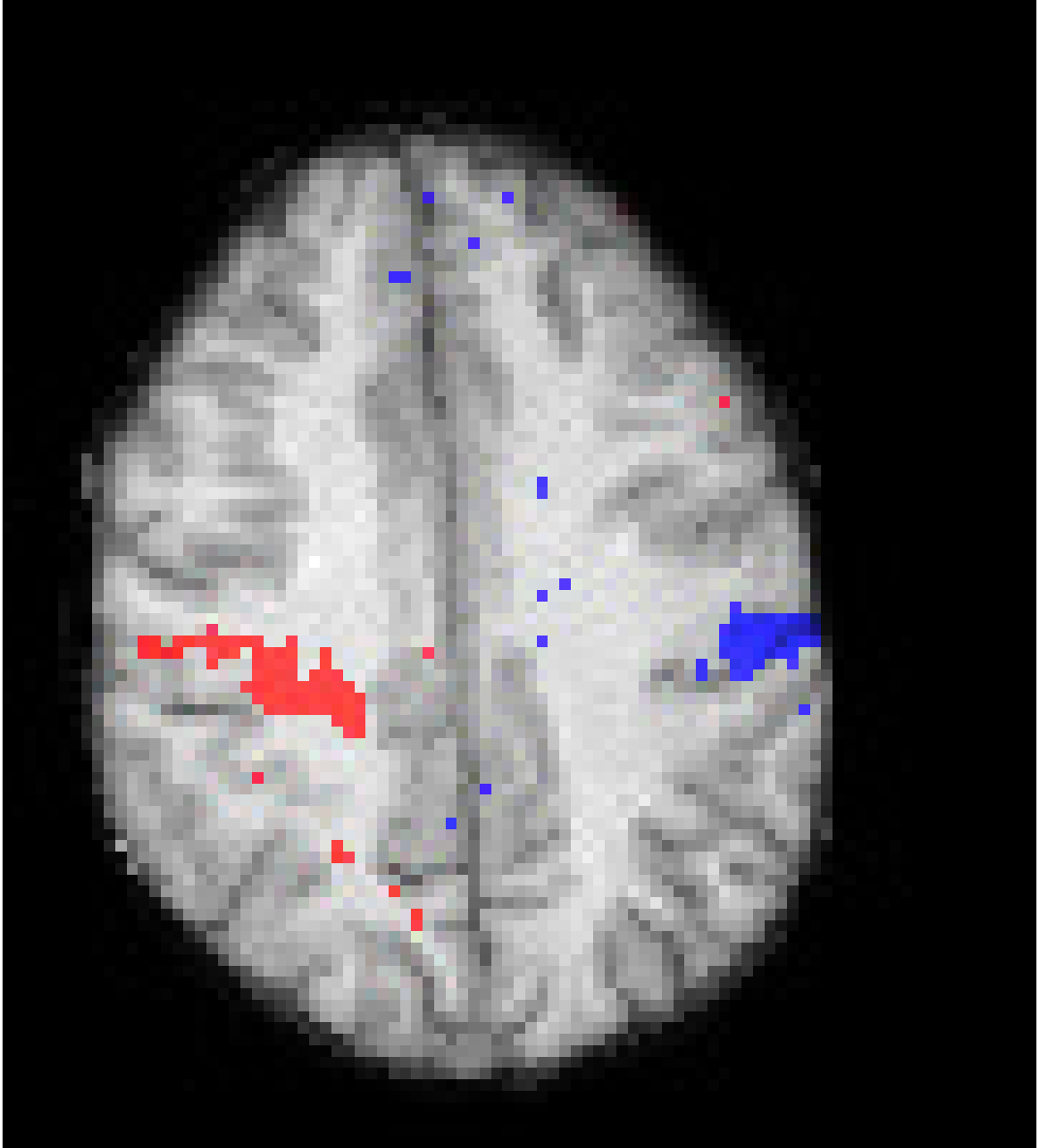}
&\includegraphics[width=\imgwidth, height=\imgheight]{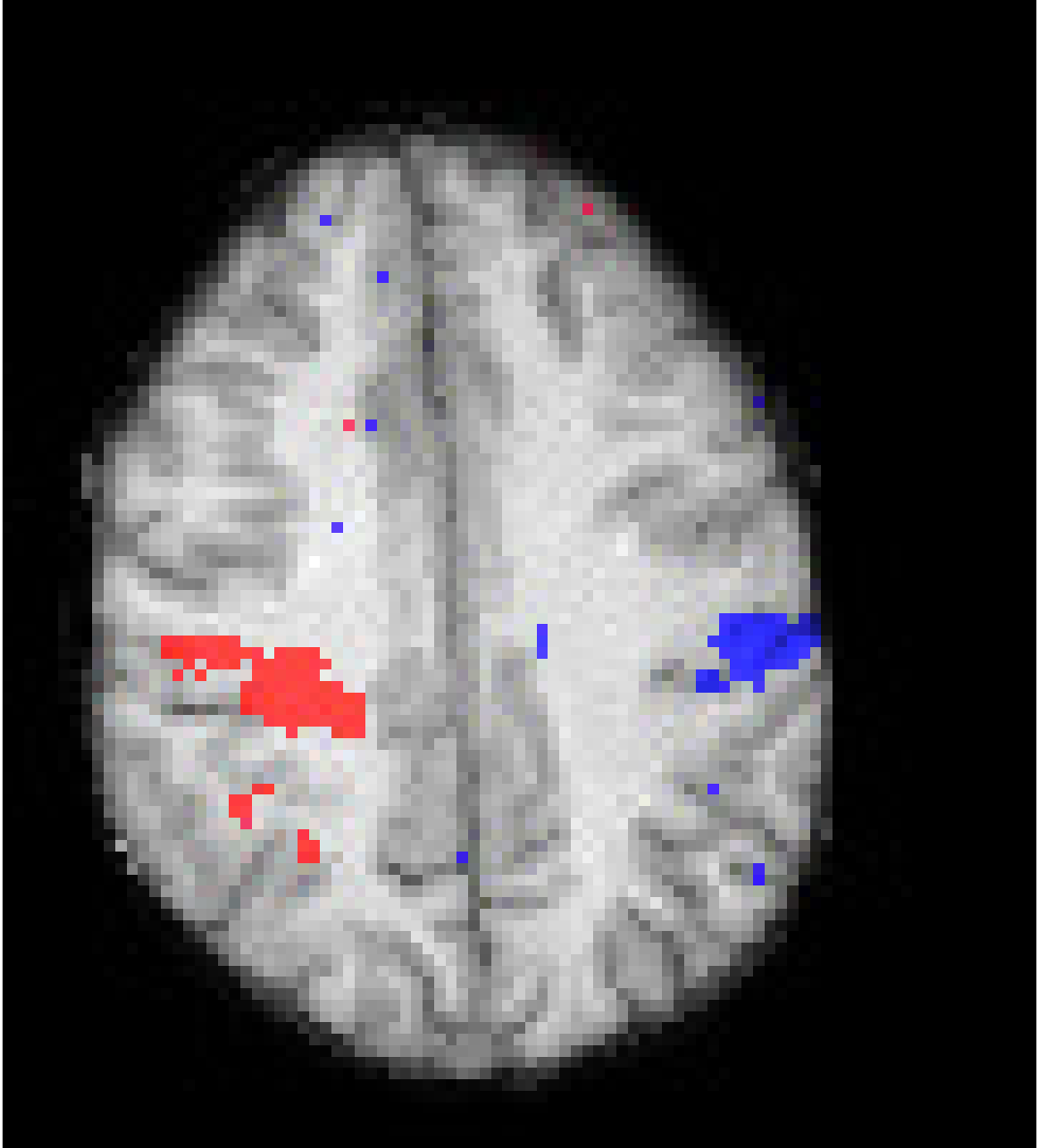}\\
\rotatebox{90}{\hspace{20pt}Slice B}
&\includegraphics[width=\imgwidth, height=\imgheight]{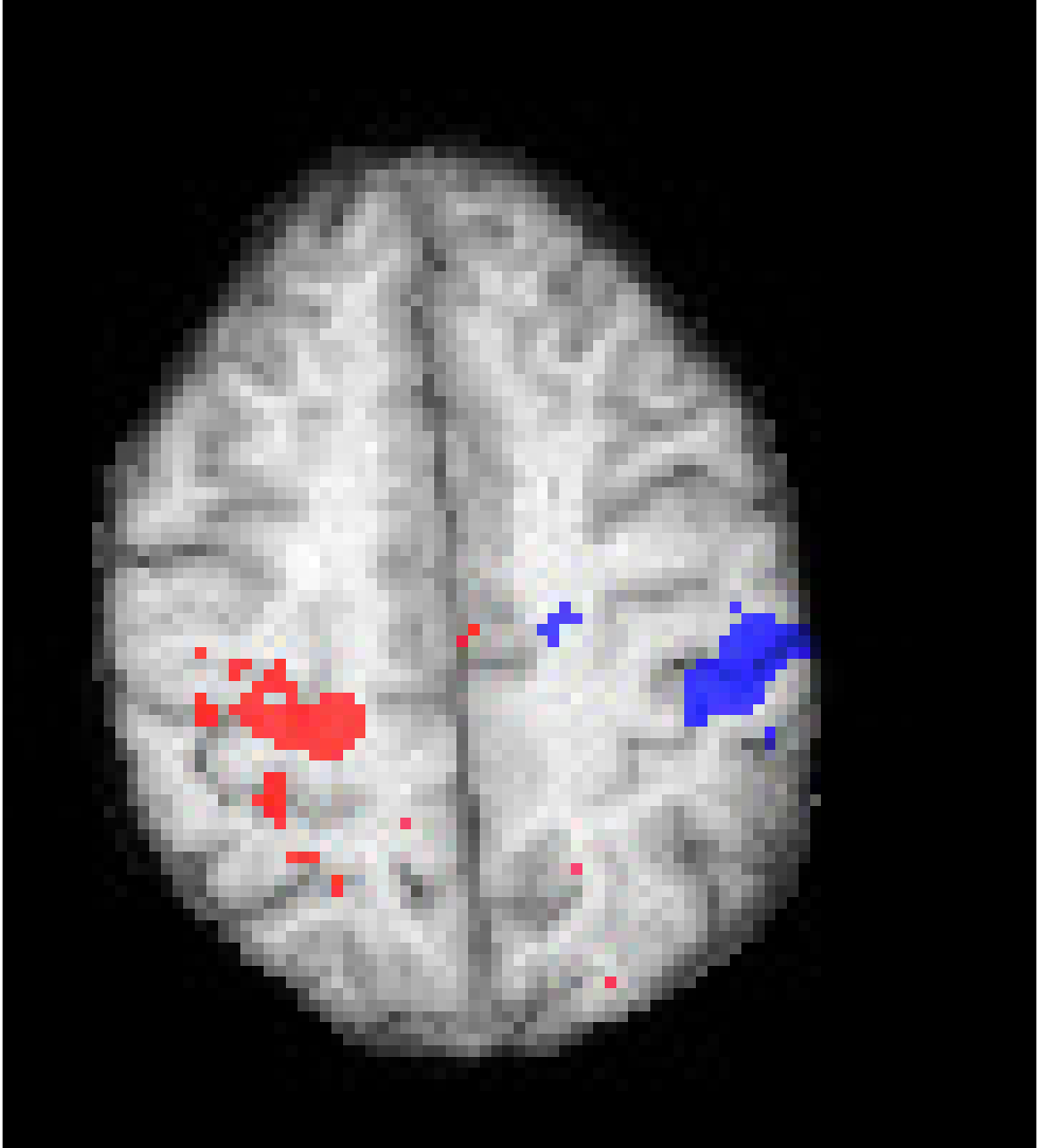}
&\includegraphics[width=\imgwidth, height=\imgheight]{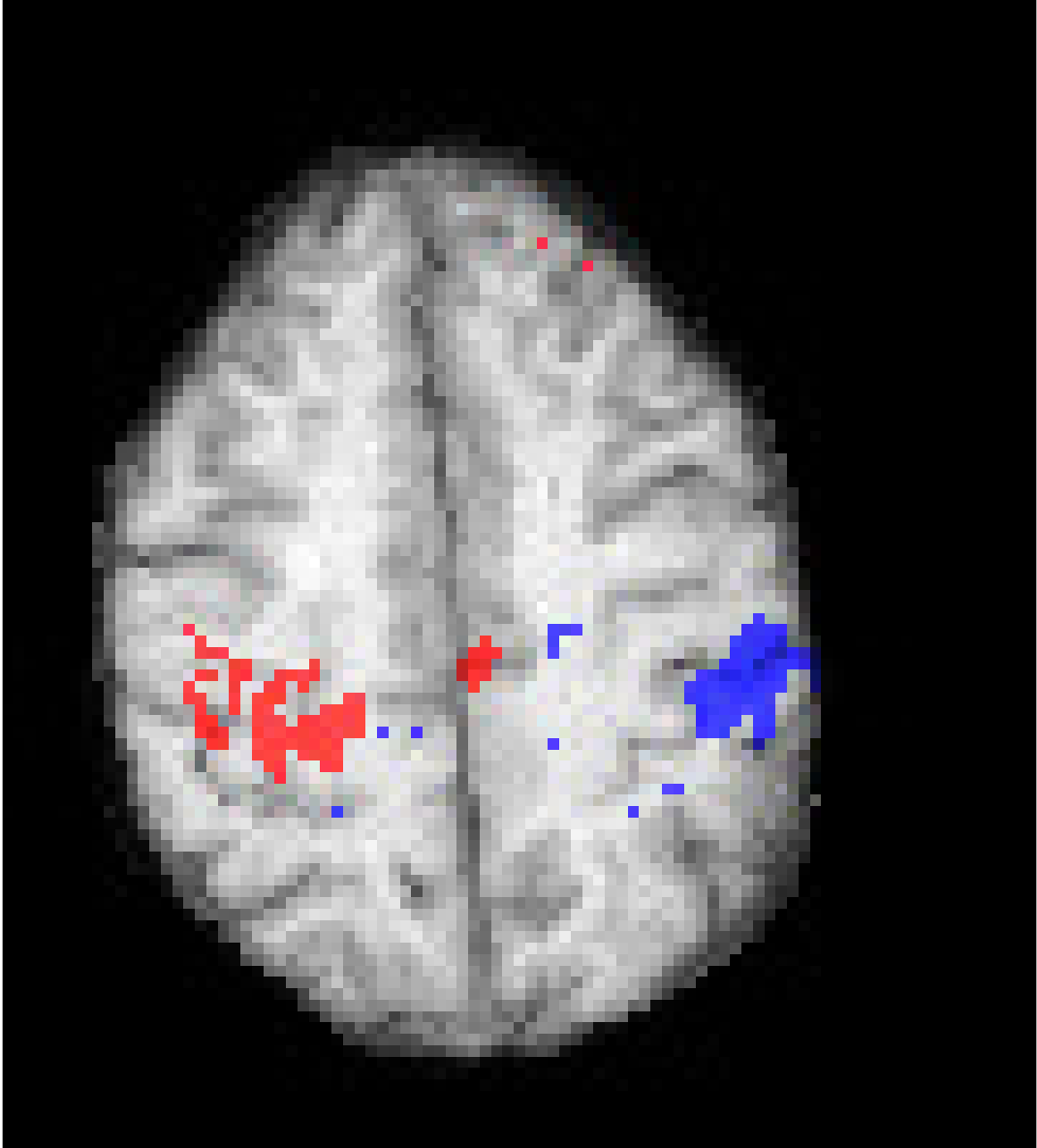}
&\includegraphics[width=\imgwidth, height=\imgheight]{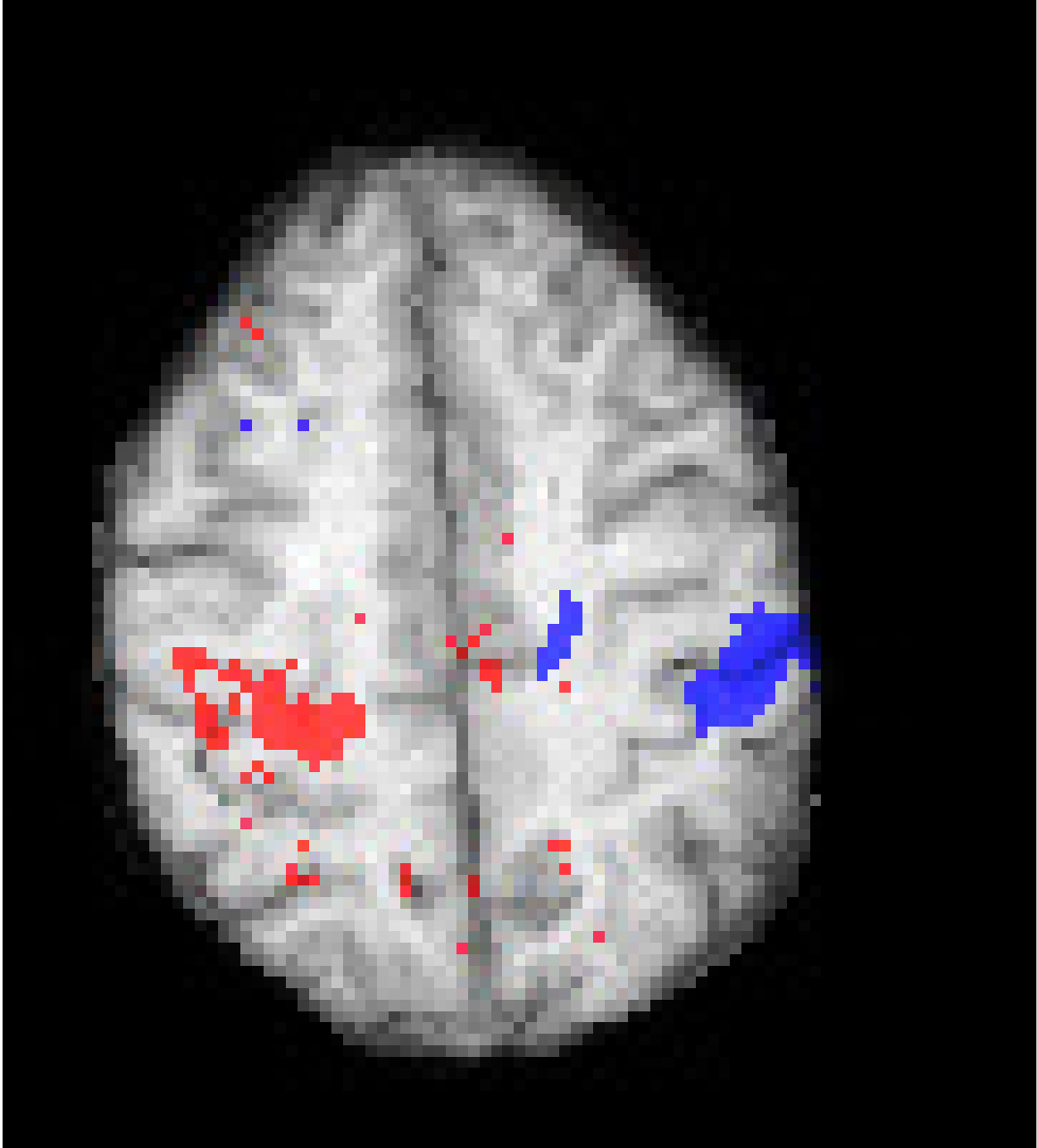}\\
\rotatebox{90}{\hspace{20pt}Slice C}
&\includegraphics[width=\imgwidth, height=\imgheight]{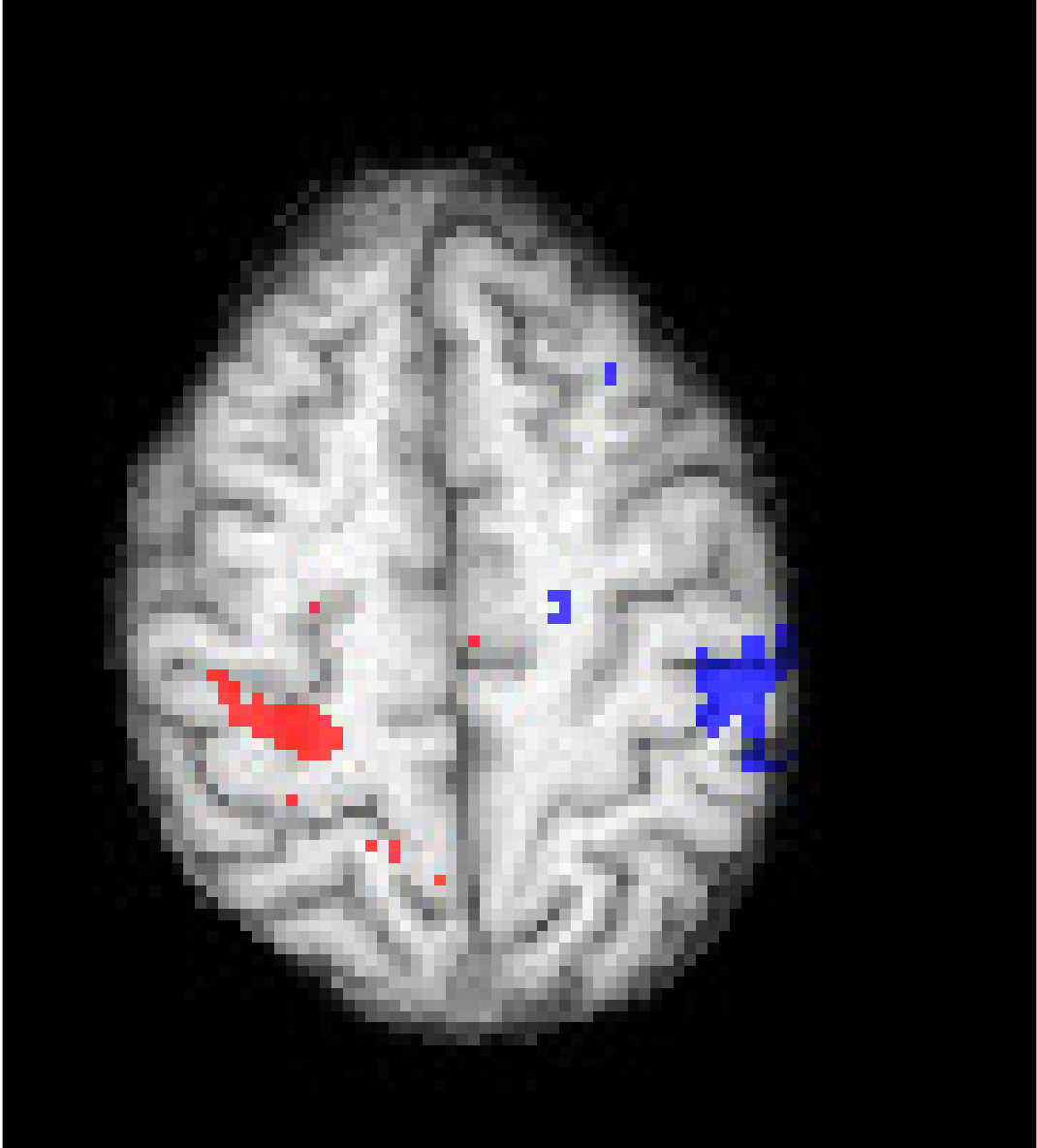} 
&\includegraphics[width=\imgwidth, height=\imgheight]{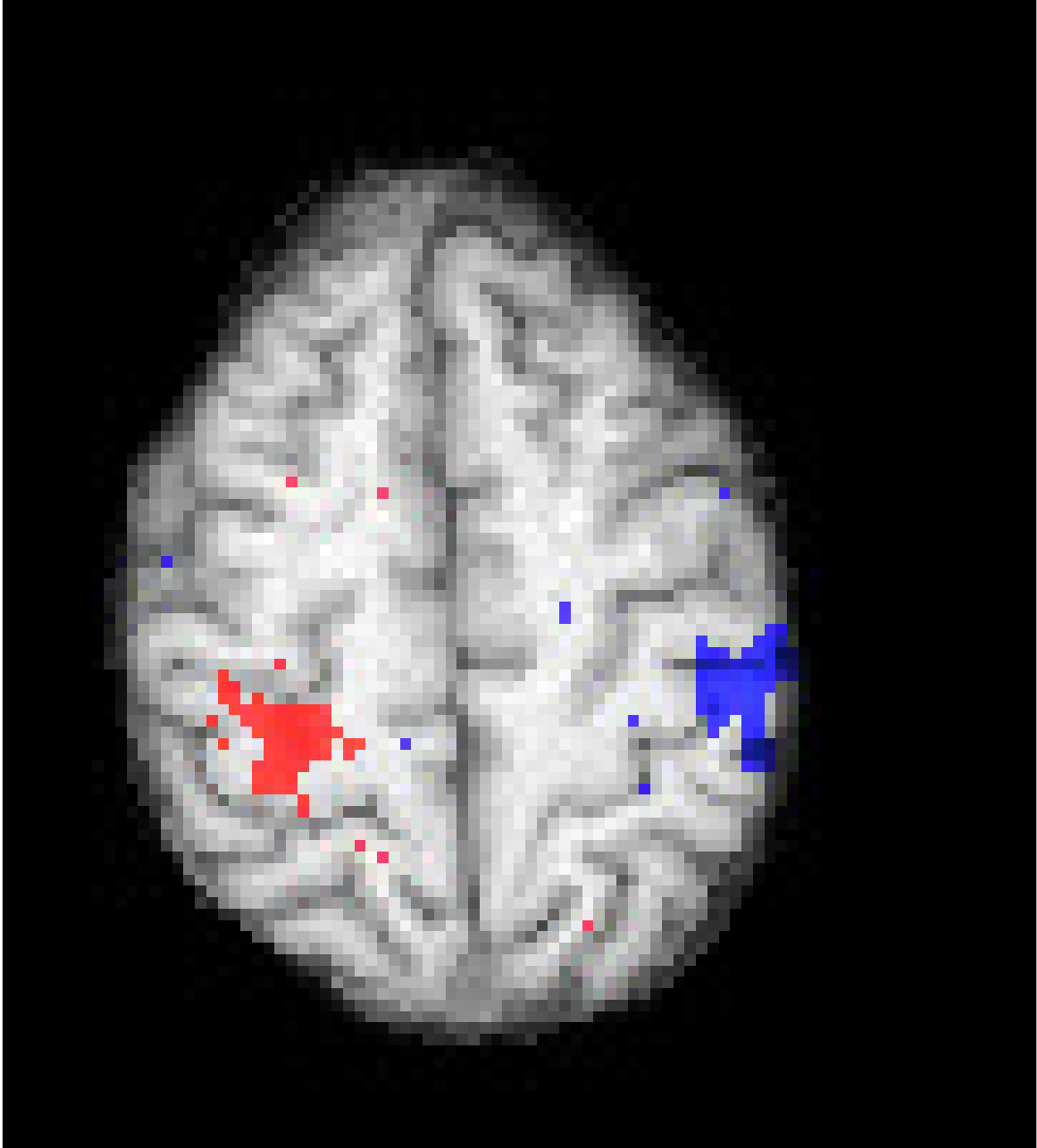}
&\includegraphics[width=\imgwidth, height=\imgheight]{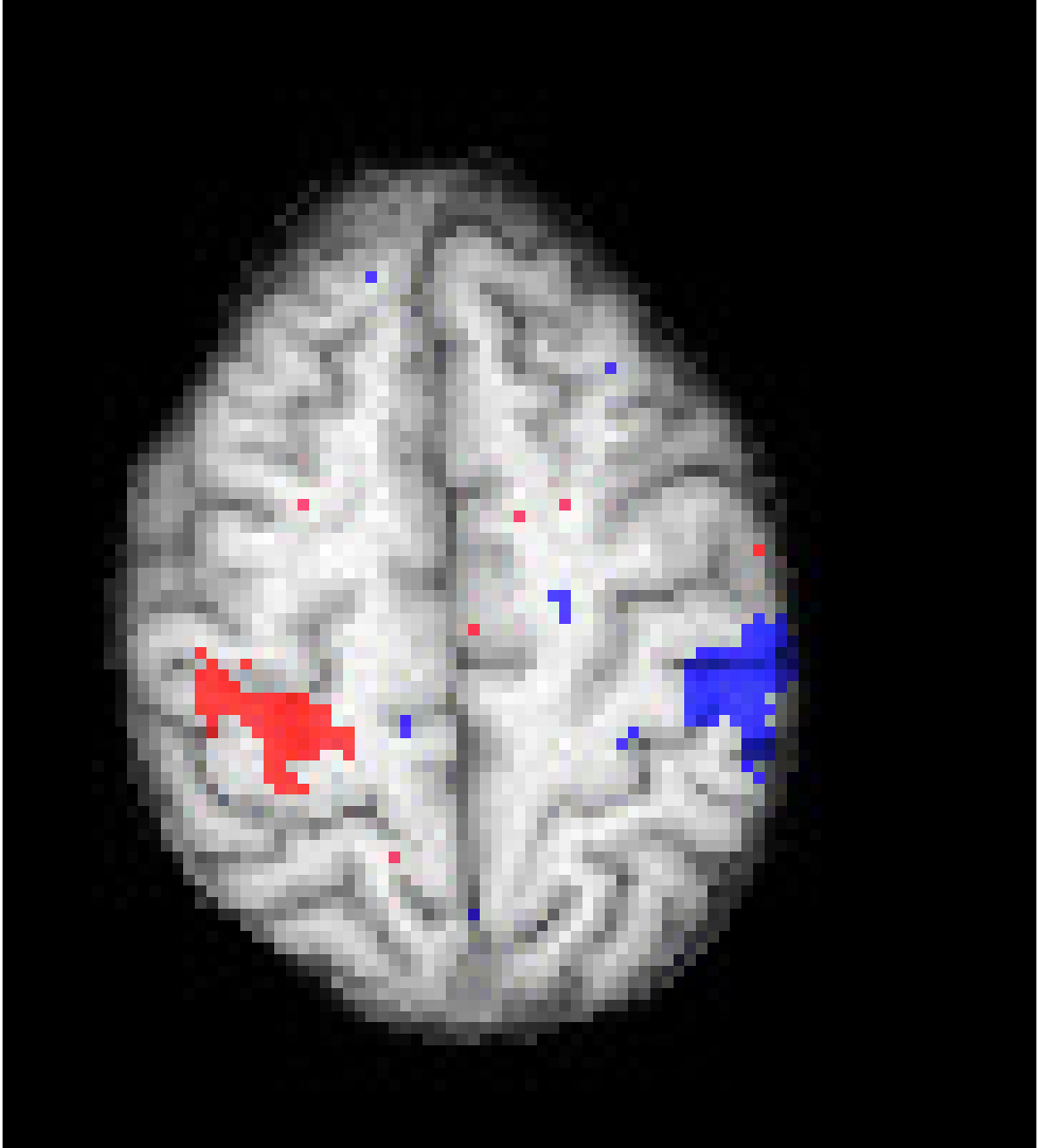}\\
&V2V&S2V&HMT
\end{array}$
}} 
\end{center}
\end{minipage}
\begin{minipage}[c]{0.49\textwidth}
\begin{center}
\subfloat[Run2 Activation Map]{{
\setlength\arraycolsep{\imgspacing}
$\begin{array}{cccc}
\rotatebox{90}{\hspace{20pt}Slice A}
&\includegraphics[width=\imgwidth, height=\imgheight]{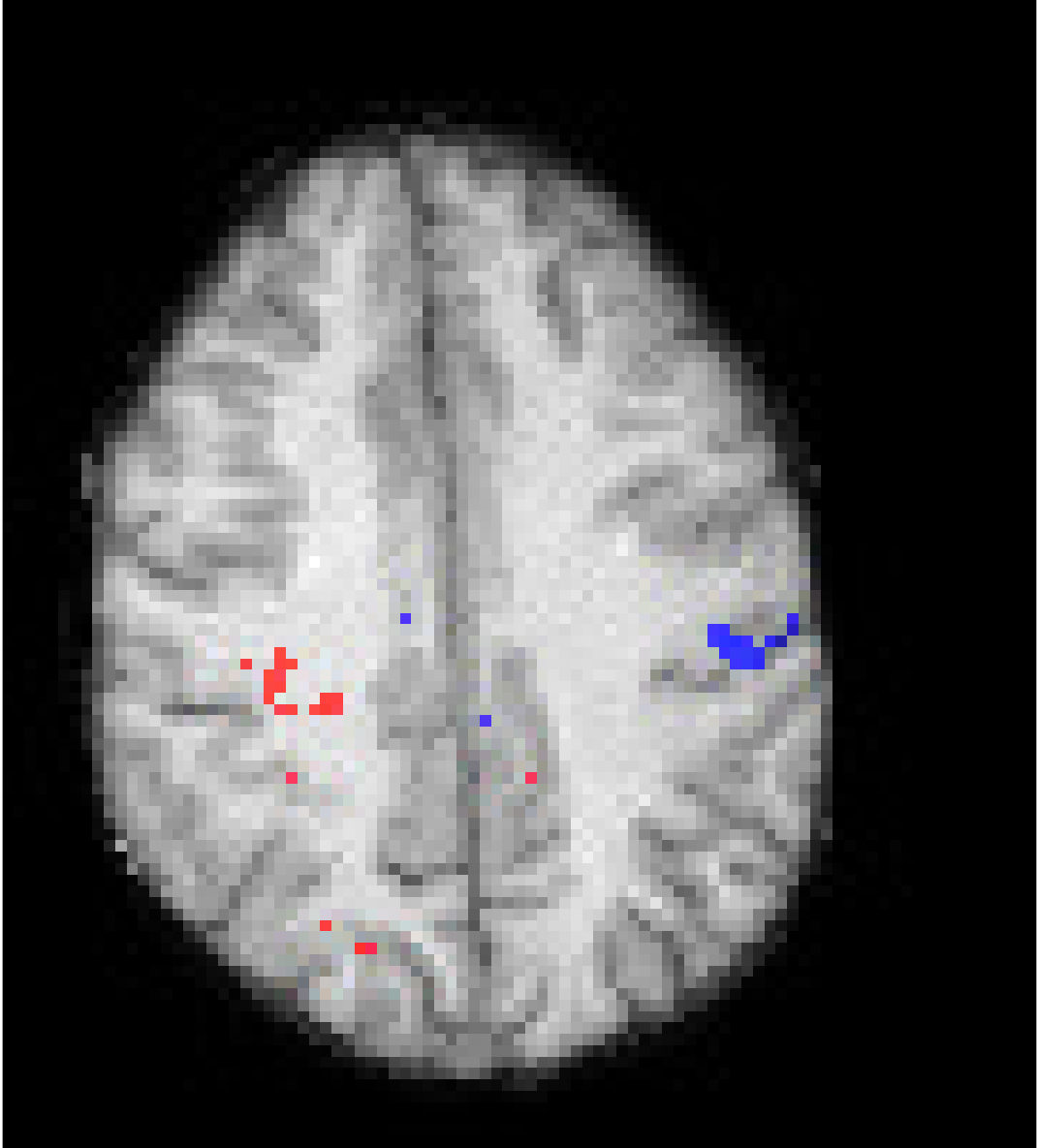}
&\includegraphics[width=\imgwidth, height=\imgheight]{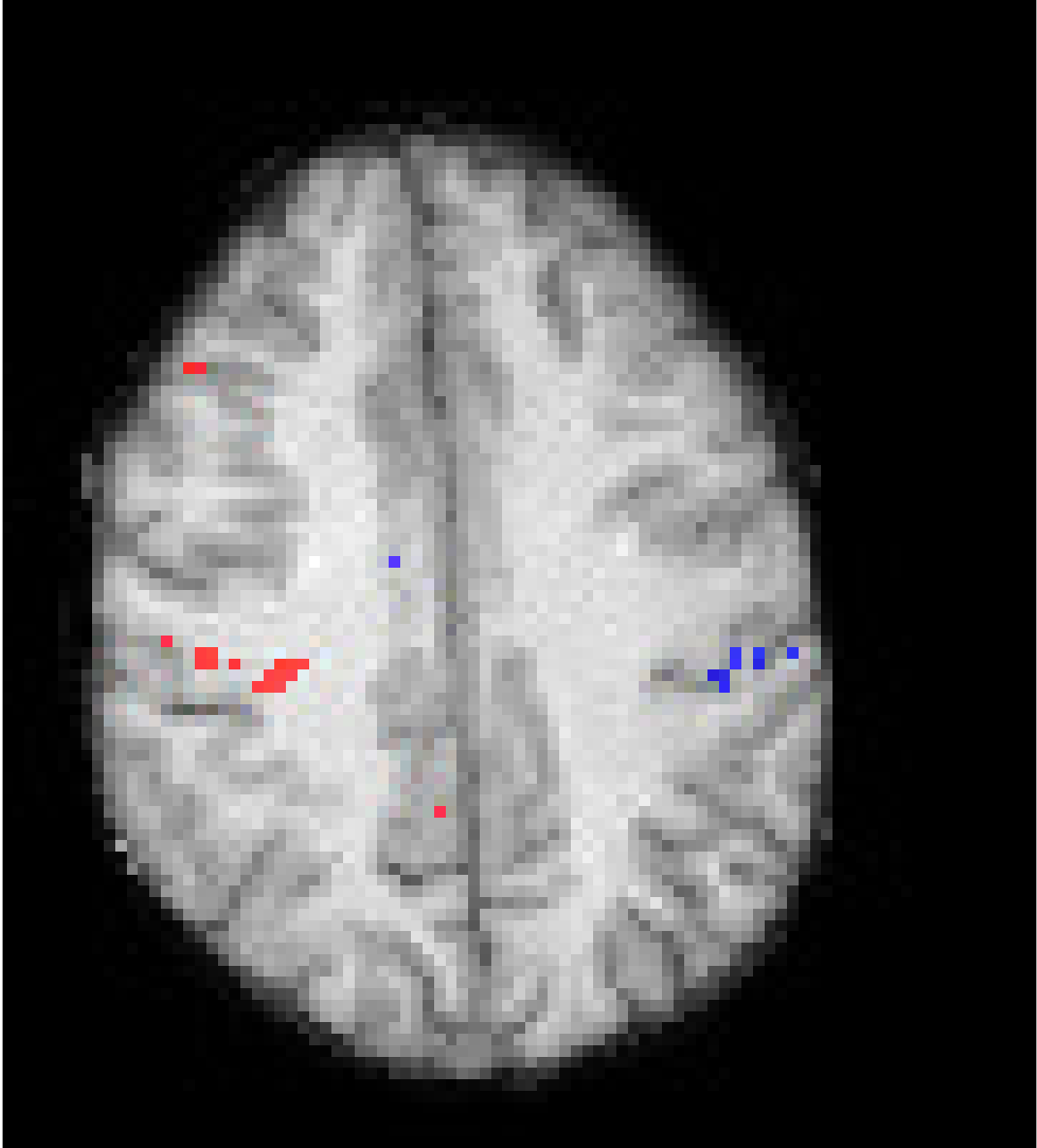}
&\includegraphics[width=\imgwidth, height=\imgheight]{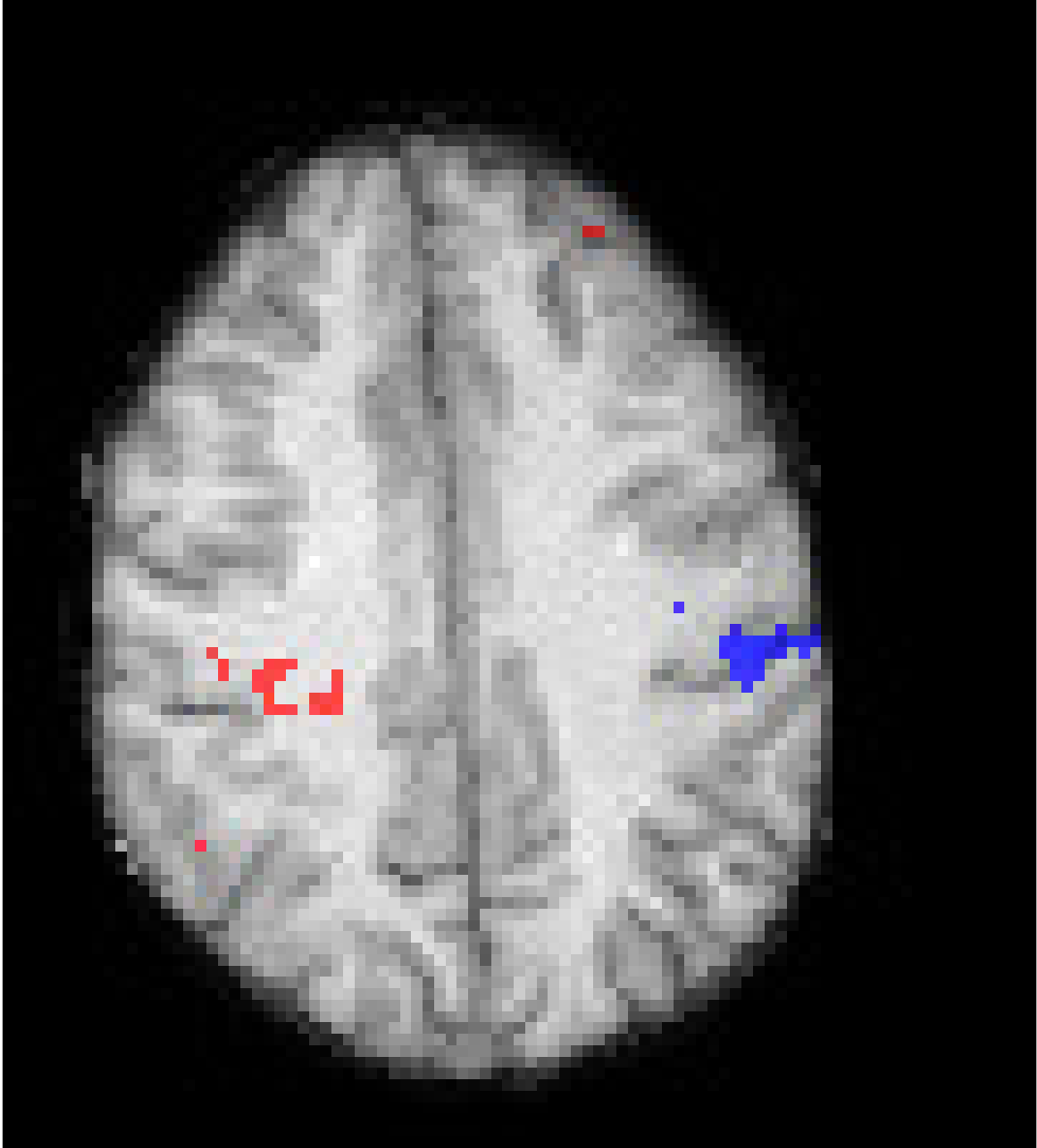}\\
\rotatebox{90}{\hspace{20pt}Slice B}
&\includegraphics[width=\imgwidth, height=\imgheight]{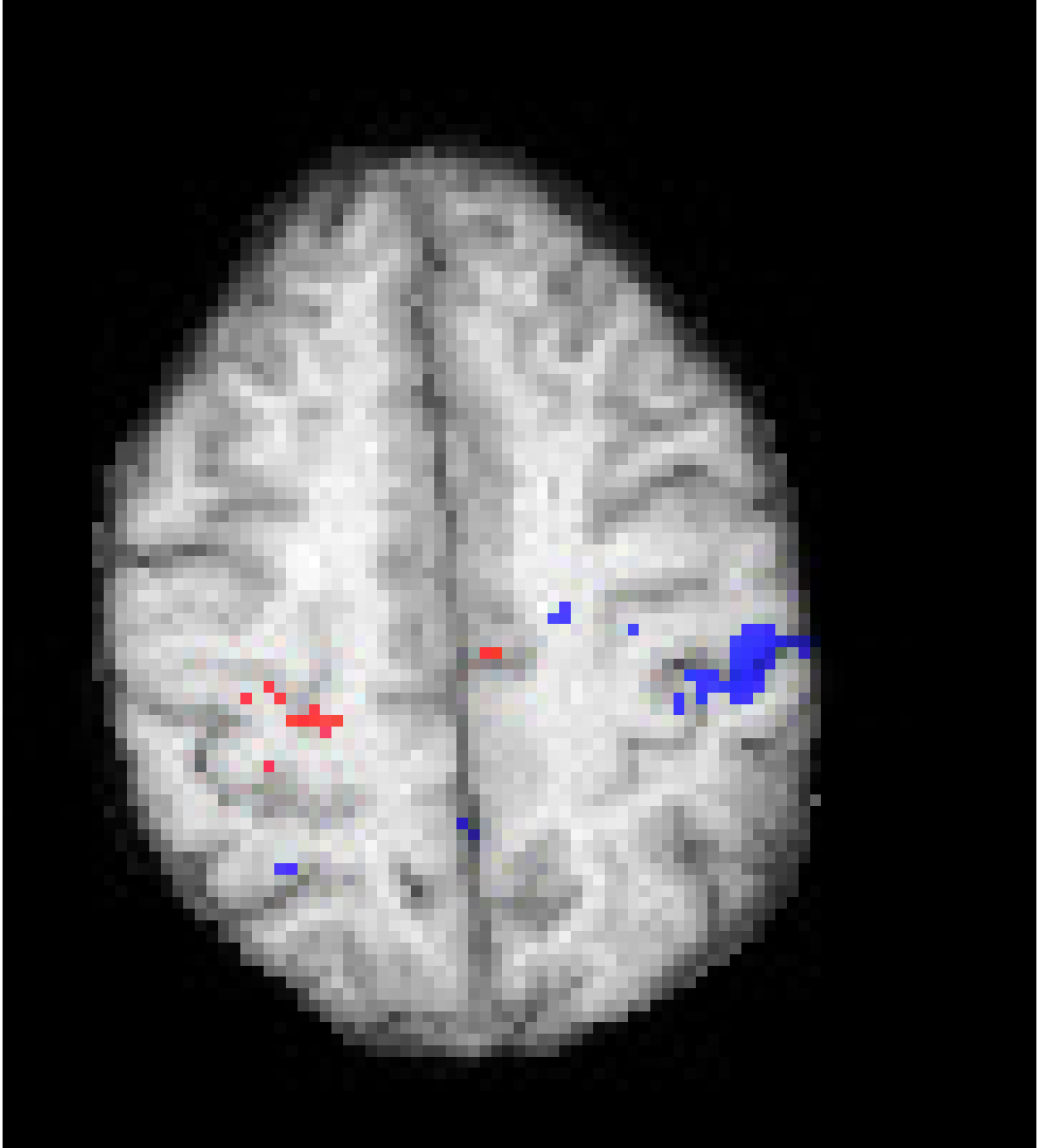}
&\includegraphics[width=\imgwidth, height=\imgheight]{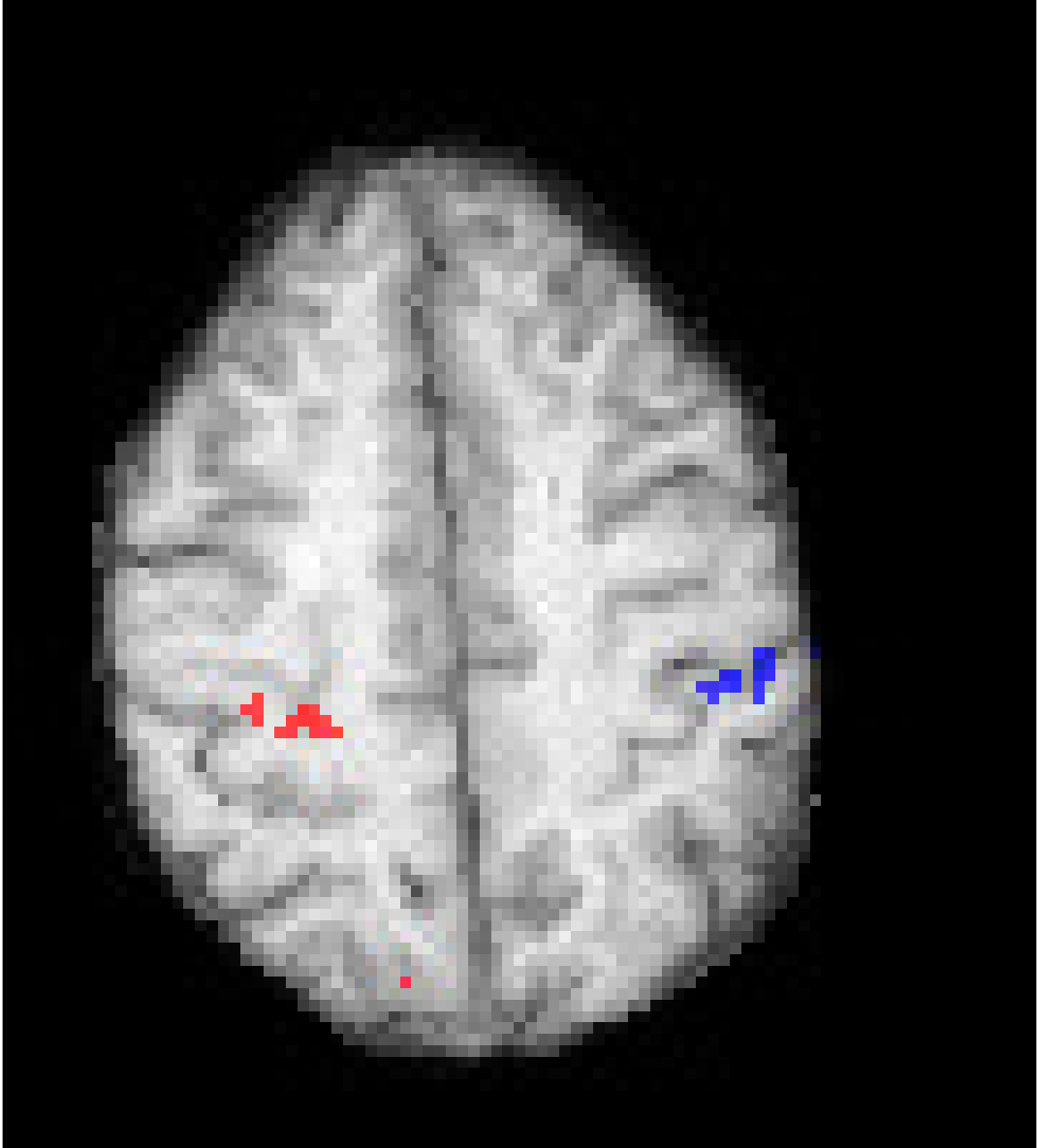}
&\includegraphics[width=\imgwidth, height=\imgheight]{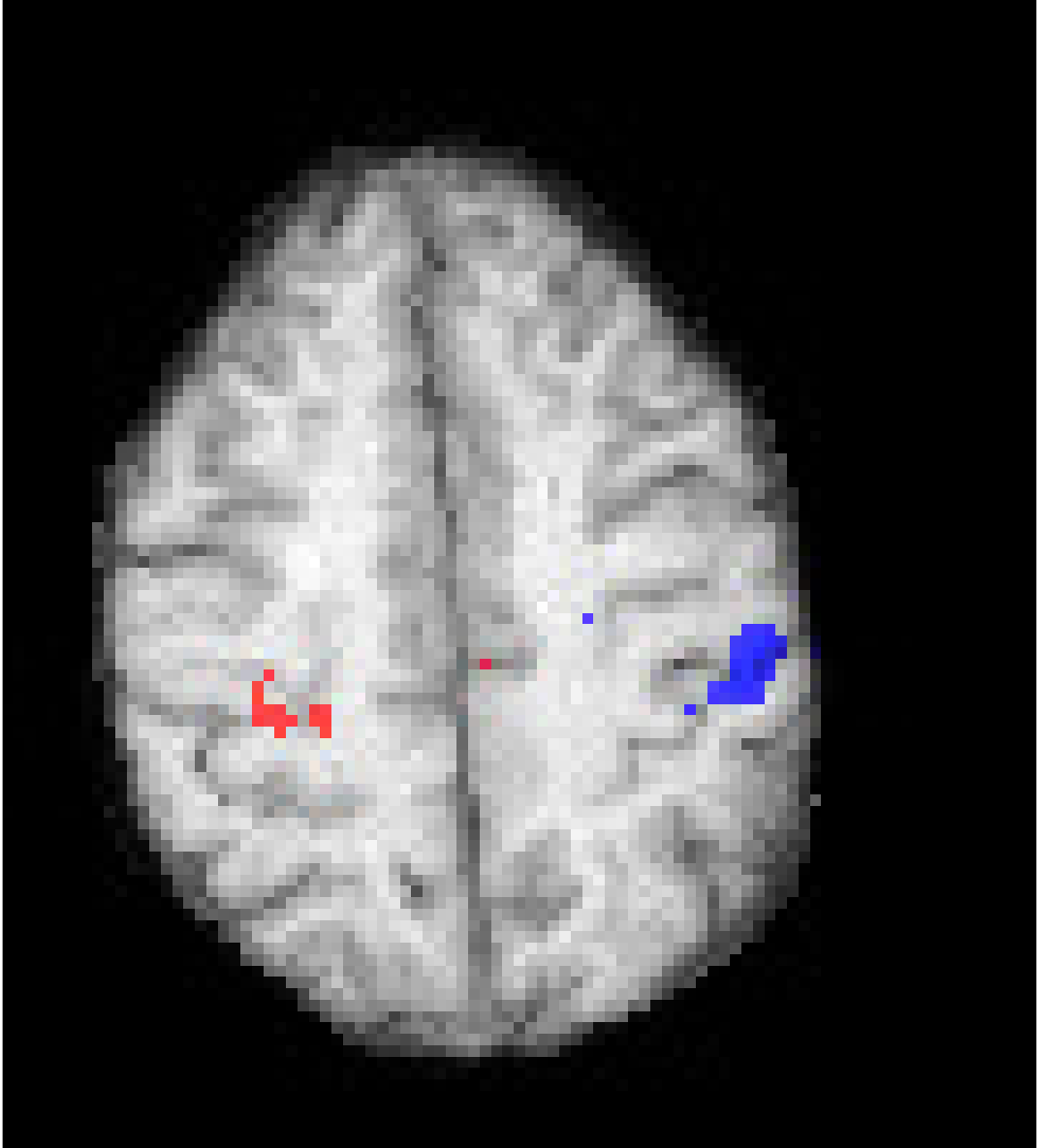}\\
\rotatebox{90}{\hspace{20pt}Slice C}
&\includegraphics[width=\imgwidth, height=\imgheight]{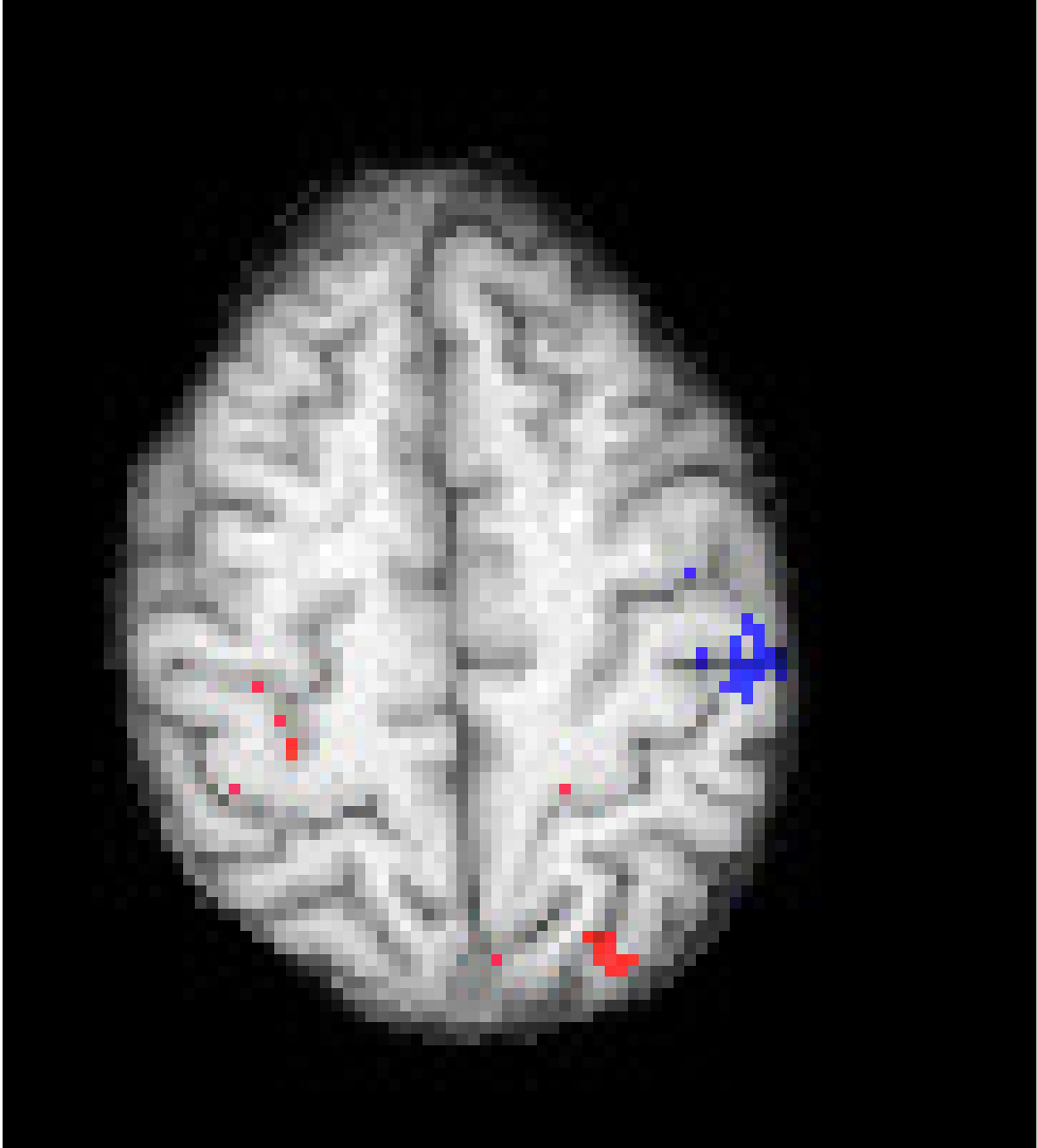}
&\includegraphics[width=\imgwidth, height=\imgheight]{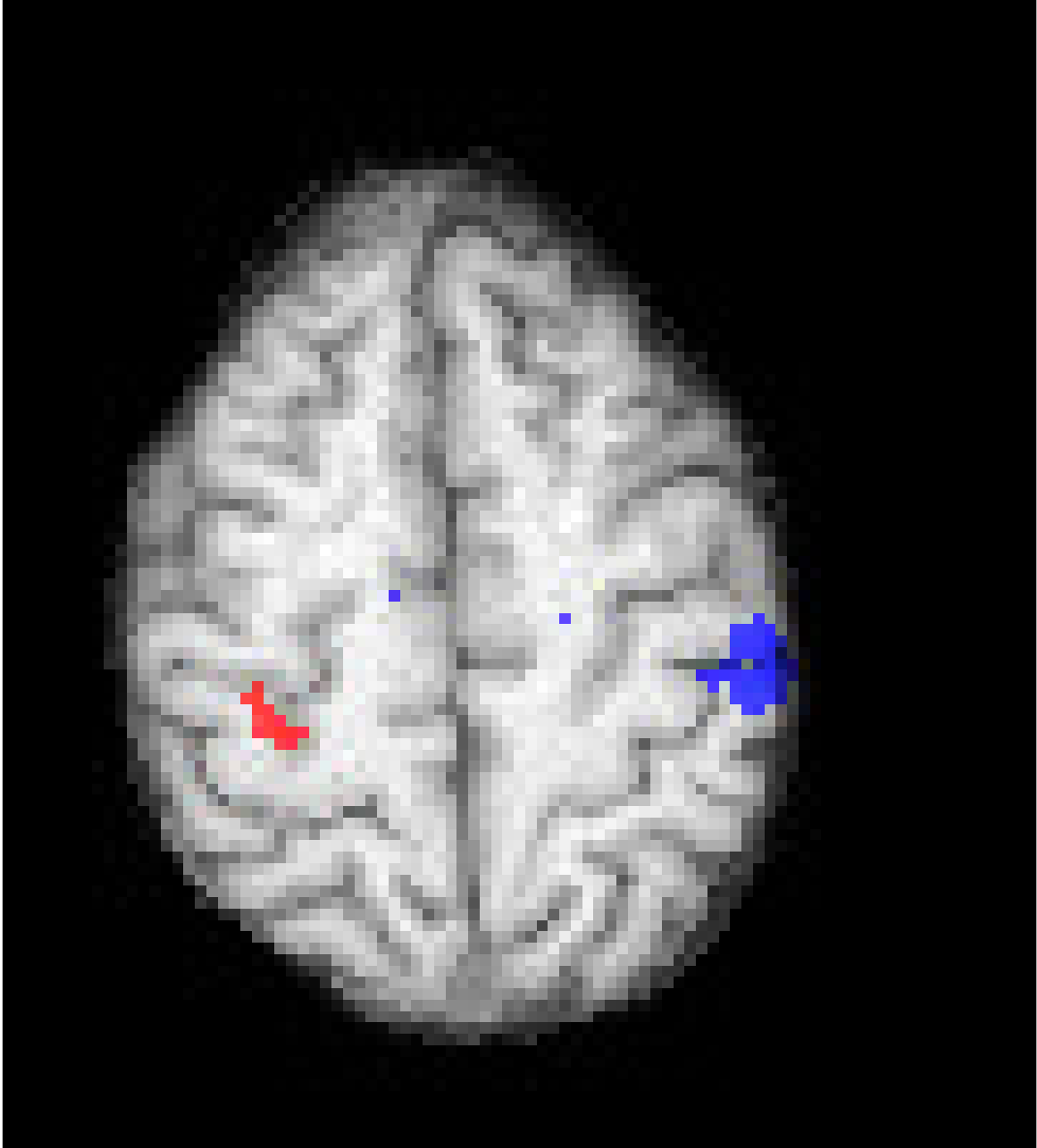}
&\includegraphics[width=\imgwidth, height=\imgheight]{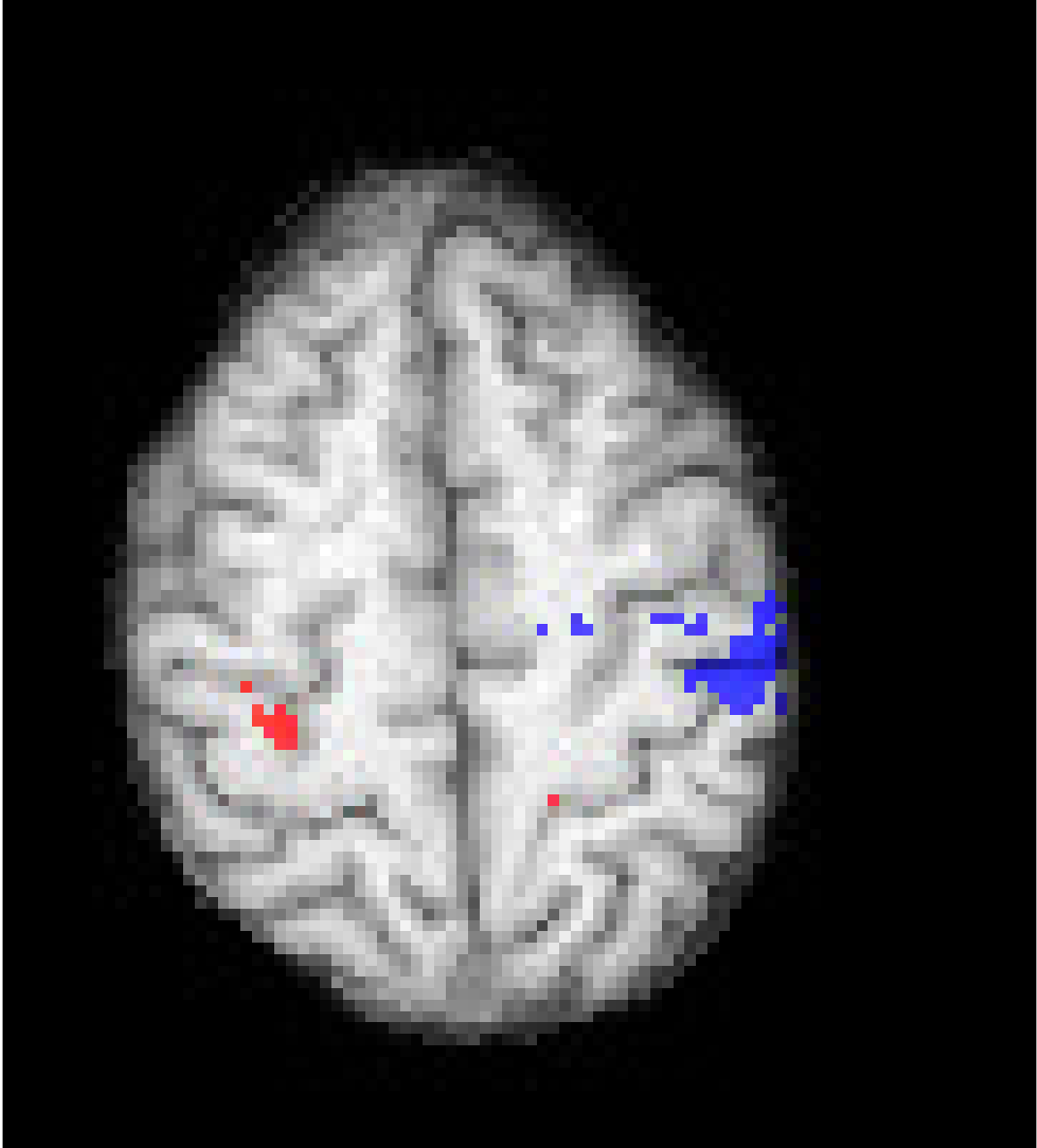}\\
&V2V&S2V&HMT
\end{array}$
}} 
\end{center}
\end{minipage}
\caption{The colorized activation maps overlaid on the anatomical MRI images for Run1 (a) and Run2 (b) datasets. The results of the three methods: (1) V2V registration; (2) S2V registration; (3) proposed HMT algorithm are listed in order from left to right column. In (a), we can see that the V2V (first column) approach produced a more dispersed set of active regions due to the inter-slice head motion. S2V (second column) produced more clustered active regions but has lots of false positive voxels scattered in the white matter. The proposed HMT (third column) generated the least dispersed active regions and had the least false positive voxels in the white matter. In (b), the activation maps from V2V and S2V (left two columns) had few and scattered active voxels due to the effect of head motion. The proposed HMT (third column) produced clean and well clustered active regions.}
\label{fig:real_actmap}
\end{figure*}

The improvement in the head tracking translates into better activation detection performance, Fig~\ref{fig:real_actmap} shows colorized activation maps overlaid on the anatomical MRI, which is used as an additional reference volume for registration. These selected slices (denoted as slice A, B, and C) displayed in different rows, show representative activated regions. Significant voxels are marked in red and blue to indicate the temporal positive and negative correlations, respectively. 

%The results of the three methods: (1) V2V; (2) S2V; (3) the proposed HMT are listed in the same order from left to right columns. 

Figure~\ref{fig:real_actmap}(a) shows the activation maps for V2V, S2V, and the proposed HMT algorithms applied to the Run1 dataset. For this easier dataset (less head motion), we can see that all methods are able to produce active regions that are near the motor cortex related to finger moves~\cite{beisteiner_finger_2001}. However, the volume-based (first column) approach produced much more spread out active regions, which may be due to small amounts of head motion. S2V (second column) did produce more clustered active regions, however, it also has some active voxels which are scattered in the white matter and are therefore likely to be false positive detections. Our proposed HMT (third column) generated active regions along the gray matter and has the least false positive voxels in the white matter. For the more challenging Run2 dataset (larger head motion), shown in Fig.~\ref{fig:real_actmap}(b), the activation maps of V2V and S2V (left two columns) have very few active voxels that are scattered across the volume. In contrast, the proposed HMT algorithm (third column) produced clean and well clustered active regions on the gray matter, which are more likely to correspond to real brain activity responses. A quantitative measure of the activation detection reliability is summarized in Table~\ref{table:ATR}. We can see that the three methods have the same level of $p_I$ values but HMT has significantly higher $p_A$, especially for the harder Run2 dataset.

\begin{table}[!htbp]
\caption{Activation Detection Reliability}
\label{table:ATR}
\begin{center}
\begin{tabular}{|c|c|c|c|c|c|c|}
\hline
\multirow{2}{*}{Method} &  \multicolumn{2}{c|}{Simulated} & \multicolumn{2}{c|}{Run1} & \multicolumn{2}{c|}{Run2} \\
\hhline{~------}
  & $p_A$ & $p_I$ & $p_A$ & $p_I$ & $p_A$ & $p_I$ \\
\hline
Truth   	&  1.000 & 0.000  &  -  & - & - & - \\
% No Corr.  &  0.004 & 0.004  & \textbf{0.711}  & 0.004  & 0.030  & 0.001 \\
V2V 		  &  0.128 & 0.003  & 0.521  & 0.003  & 0.047  & 0.001 \\
S2V		    &  0.248 & 0.002  & 0.614  & 0.003  & 0.048  & 0.001 \\
HMT   		&  \textbf{0.662} & 0.003  & \textbf{0.623}  & 0.003  & \textbf{0.087}  & 0.002 \\
\hline
\end{tabular}
\\
\smallskip
\justify
The proposed HMT algorithm attains significantly higher $p_A$, especially for Run2 dataset, while keeps the same level of $p_I$ compared to the other motion compensation algorithms (V2V, S2V).
\end{center}
\end{table}

\section{Conclusion}
\label{sec:conclusion}
In this work, we have proposed a head motion tracking (HMT) algorithm that uses an image registration objective function combined with a Gaussian particle filter to couple motion estimates from successive EPI slices, resulting in improved performance. Due to the fact that the proposed algorithm utilizes the information from consecutive slices in the fMRI scan volume, it effectively combines the bias reduction properties of the S2V approach and the variance reduction properties of the V2V approach.

Evaluation based on synthetic data demonstrated that the proposed HMT algorithm can significantly improve accuracy over the volume-to-volume and slice-to-volume approaches in terms of motion parameter estimation and activation detection accuracy. Using real human experimental data we demonstrated that the proposed algorithm is able to produce more stable estimates of head motion and brain activation maps. Unlike previous approaches to head motion compensation, the activation maps of the HMT produce more reliable active regions even when the head motion is large during the fMRI scan.

Improvements in robustness and accuracy of the proposed HMT algorithm may permit scientists to analyze more complex brain activation patterns. This can be especially beneficial for experiments that involve a wider spatial distribution activation regions, and are more likely to have motion artifacts, e.g., in working memory or speech experiments. Furthermore, our HMT approach might allow fMRI to be reliably applied to patients having significant motion disorders, e.g., Parkinson's disease, who currently do not benefit from fMRI examinations. 

\appendix[Particle Weights Evaluation]
\label{append:particle_weight_evaluation}
The particle weights are evaluated through the quasi-likelihood function $p(\bS_t|\btheta_t)$. The quasi-likelihood function should have two properties: (1) It is monotonically increasing with the image similarity $\bbM(\bS_t, T_\btheta^*(V_{\mathrm{anat}}))$; (2) The weighted particles are distributed approximately to multivariate Gaussian. To satisfy the two properties, we propose to use a histogram equalization approach to evaluate the particle weights. The multivariate Gaussian density is shown below:

\begin{equation}
\label{eq:Gaussian_density}
f(\bx)=\frac{1}{\sqrt{(2\pi)^d|\bSigma|}}\exp{-\frac{1}{2}(\bx-\bmu)^T\bSigma^{-1}(\bx-\bmu)}.
\end{equation}

The goal here is to find the distribution of $z=f(\bx)$ where $\bx$ is the random variable following (\ref{eq:Gaussian_density}). Let $g_Z(z)$ denote the density of $z$. We can equalize the histogram of image similarity to $g_Z(z)$ to obtain the particle weights. 

Without loss of generality and for simplicity, in the following derivation, we assume the covariance to be identity matrix and $\bmu=\mathbf 0$. The density function and its inverse can be re-written as:
\begin{equation}
\label{eq:Gaussian_density_simplified}
\begin{split}
h(r)&=\frac{1}{\sqrt{(2\pi)^d}}\exp{-\frac{1}{2}r^2}, \\
h^{-1}(z)&=\sqrt{-2\log{\left(\sqrt{(2\pi)^d}z\right)}},~z\in (0,\sqrt{(2\pi)^{-d}}],
% h^{-1}(z)&=\sqrt{-2\log z-d\log{2\pi}},~z\in (0,\sqrt{(2\pi)^{-d}}]
\end{split}
\end{equation} 
where $r=\|\bx\|$. Define $G_Z(z)$ as cumulative density function of $g_Z(z)$ (i.e. $G_Z(z)=p(\{f(\bx) \le z\})$), where $\bx$ is the random variable following the multivariate Gaussian density (\ref{eq:Gaussian_density}). According to the spherical symmetry, $G(z)$ has the following form by integration along the radial direction:
\begin{equation}
\label{eq:Gy_func}
\begin{split}
G(z)&=1-p(\{f(\bx)\ge z\})\\
&=1-\int_0^{h^{-1}(z)}\mS_{d-1}u^{d-1} h(u)du \\
&=1-\mS_{d-1}\left(H^*(h^{-1}(z)) - H^*(0)\right),
\end{split}
\end{equation}
\begin{equation}
\label{eq:H_integral}
H^*(u) = \int_{-\infty}^uv^{d-1}h(v)dv,
\end{equation}
where $\mS_{d-1}$ is the surface area of unit $(d-1)$-sphere, e.g., $\mS_0=2, \mS_1=2\pi$. To obtain $g(z)$, we need to take the derivative of $G(z)$ with respect to $z$:
\begin{equation}
\label{eq:derivative_of_Gz}
\begin{split}
g(z) &= \frac{dG(z)}{dz}=-\mS_{d-1}\frac{dH^*(h^{-1}(z))}{dz} \\
&= -\mS_{d-1}\frac{H^*(h^{-1}(z))}{dh^{-1}(z)}\frac{dh^{-1}(z)}{dz},
\end{split}
\end{equation}

\begin{equation}
\label{eq:derivative_of_Hz}
\begin{split}
\frac{H^*(h^{-1}(z))}{dh^{-1}(z)}&=z\left(h^{-1}(z)\right)^{d-1},
\end{split}
\end{equation}

\begin{equation}
\label{eq:derivative_of_hinvz}
\frac{dh^{-1}(z)}{dz} = -\frac{1}{zh^{-1}(z)}.
\end{equation}

By substituting (\ref{eq:derivative_of_Hz})(\ref{eq:derivative_of_hinvz}) into (\ref{eq:derivative_of_Gz}), we have:
\begin{equation}
\label{eq:gz}
\begin{split}
g_Z(z)&=\mS_{d-1}\left(-2\log{\left(\sqrt{(2\pi)^d}z\right)}\right)^{(d-2)/2} \\
&=\frac{d\pi^{d/2}}{\Gamma(\frac{d}{2}+1)}\left(-2\log{\left(\sqrt{(2\pi)^d}z\right)}\right)^{(d-2)/2}.
\end{split}
\end{equation}

\begin{figure}[ht]
\begin{center}
\subfloat[]{{\includegraphics[width=0.47\columnwidth]{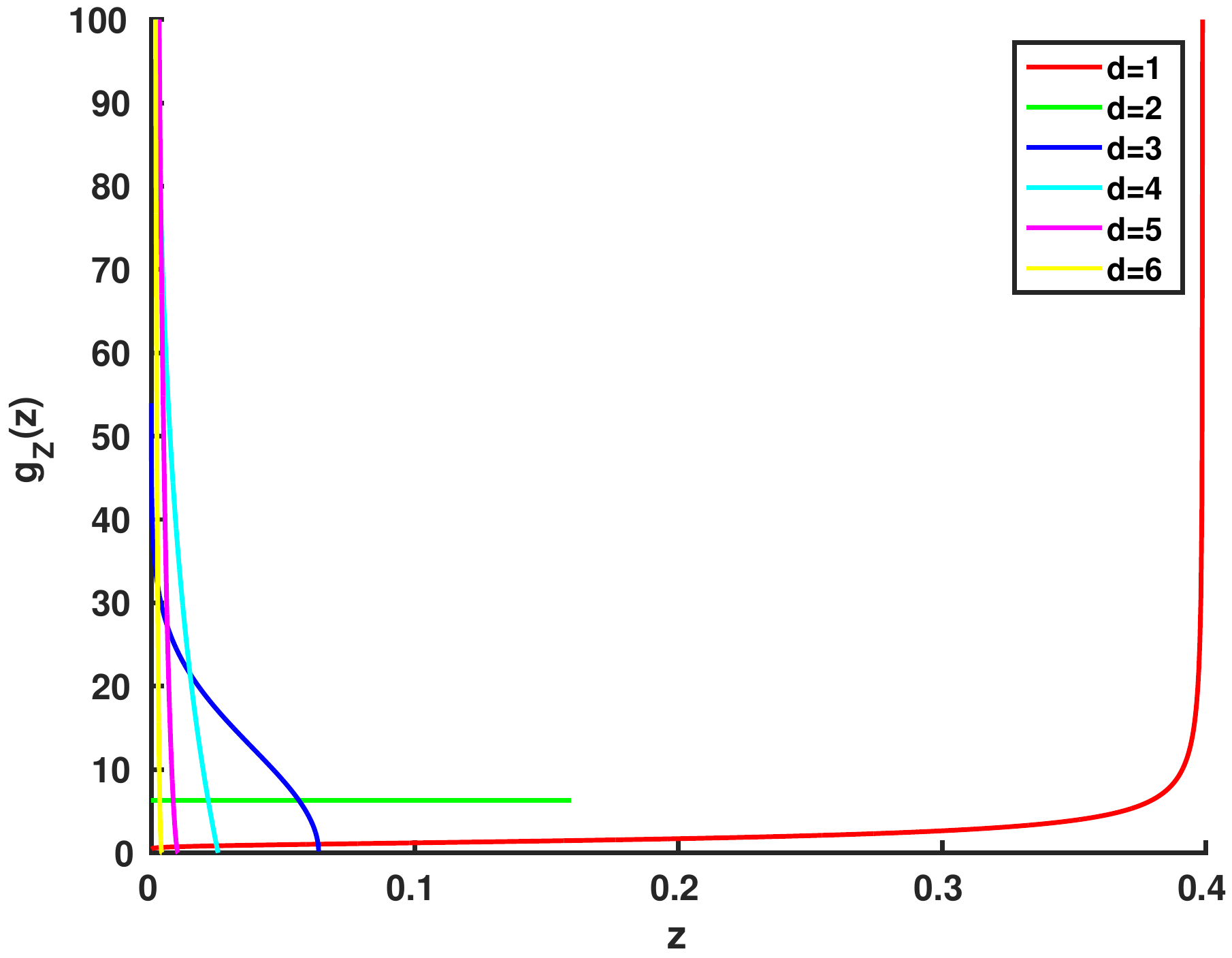}}}
\subfloat[]{{\includegraphics[width=0.47\columnwidth]{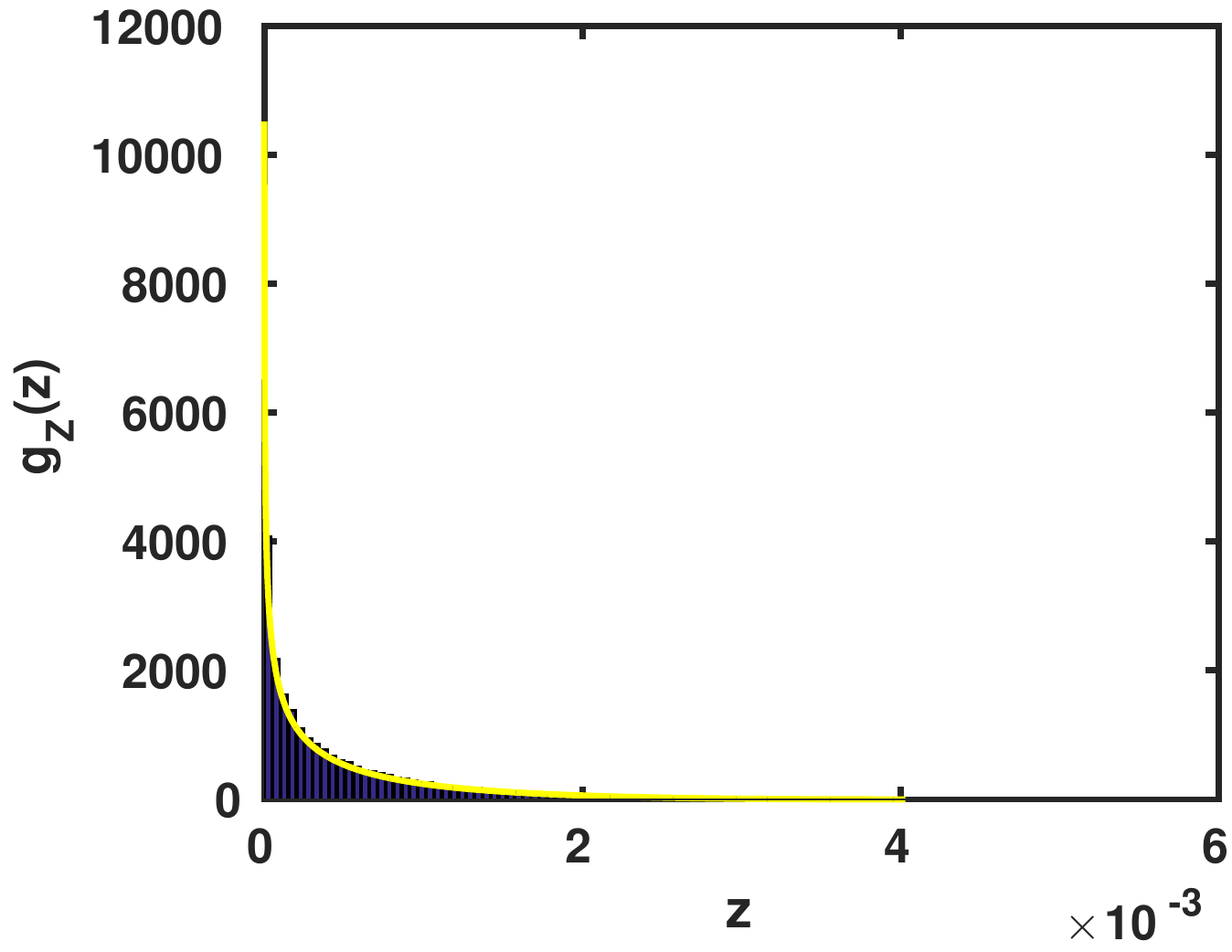}}}
\end{center}
\caption{(a) shows the density $g_Z(z)$ for different $d$. (b) shows the simulated histogram compared with theoretical $g_Z(z)$ for $d=6$.}
\label{fig:GauHist}
\end{figure}

Figure~\ref{fig:GauHist}(a) shows the density $g_Z(z)$ for different dimension $d$. Notice that in this paper, the multivariate Gaussian is used to model the rigid body head motion which has $6$ dimensions and therefore $g_Z(z)$ has the following form:
\begin{equation}
\label{eq:gz_d6}
g_Z(z)=\pi^3\left(-2\log{(2\pi)^3z}\right)^2,~z\in (0,(2\pi)^{-3}].
\end{equation}

Figure~\ref{fig:GauHist}(b) plots (\ref{eq:gz_d6}) with the simulated histogram. The histogram of image similarity is equalized to (\ref{eq:gz_d6}) to obtain the weights of each particle.

% \section*{Acknowledgment}
% The authors would like to thank...

% Can use something like this to put references on a page
% by themselves when using endfloat and the captionsoff option.
\ifCLASSOPTIONcaptionsoff
  \newpage
\fi

% references section
\bibliographystyle{IEEEtran}
\bibliography{TMI2015}

% Generated by IEEEtran.bst, version: 1.12 (2007/01/11)
\begin{thebibliography}{10}
\providecommand{\url}[1]{#1}
\csname url@samestyle\endcsname
\providecommand{\newblock}{\relax}
\providecommand{\bibinfo}[2]{#2}
\providecommand{\BIBentrySTDinterwordspacing}{\spaceskip=0pt\relax}
\providecommand{\BIBentryALTinterwordstretchfactor}{4}
\providecommand{\BIBentryALTinterwordspacing}{\spaceskip=\fontdimen2\font plus
\BIBentryALTinterwordstretchfactor\fontdimen3\font minus
  \fontdimen4\font\relax}
\providecommand{\BIBforeignlanguage}[2]{{%
\expandafter\ifx\csname l@#1\endcsname\relax
\typeout{** WARNING: IEEEtran.bst: No hyphenation pattern has been}%
\typeout{** loaded for the language `#1'. Using the pattern for}%
\typeout{** the default language instead.}%
\else
\language=\csname l@#1\endcsname
\fi
#2}}
\providecommand{\BIBdecl}{\relax}
\BIBdecl

\bibitem{turner_functional_1998}
R.~Turner, A.~Howseman, G.~E. Rees, O.~Josephs, and K.~Friston, ``Functional
  magnetic resonance imaging of the human brain: data acquisition and
  analysis,'' \emph{Experimental Brain Research}, vol. 123, no. 1-2, pp. 5--12,
  1998.

\bibitem{zaitsev_magnetic_2006}
M.~Zaitsev, C.~Dold, G.~Sakas, J.~Hennig, and O.~Speck,
  ``\BIBforeignlanguage{eng}{Magnetic resonance imaging of freely moving
  objects: prospective real-time motion correction using an external optical
  motion tracking system},'' \emph{\BIBforeignlanguage{eng}{NeuroImage}},
  vol.~31, no.~3, pp. 1038--1050, Jul. 2006.

\bibitem{qin_prospective_2009}
L.~Qin, P.~van Gelderen, J.~A. Derbyshire, F.~Jin, J.~Lee, J.~A. de~Zwart,
  Y.~Tao, and J.~H. Duyn, ``\BIBforeignlanguage{eng}{Prospective head-movement
  correction for high-resolution {MRI} using an in-bore optical tracking
  system},'' \emph{\BIBforeignlanguage{eng}{Magnetic resonance in medicine:
  official journal of the Society of Magnetic Resonance in Medicine / Society
  of Magnetic Resonance in Medicine}}, vol.~62, no.~4, pp. 924--934, Oct. 2009.

\bibitem{ooi_prospective_2009}
M.~B. Ooi, S.~Krueger, W.~J. Thomas, S.~V. Swaminathan, and T.~R. Brown,
  ``\BIBforeignlanguage{eng}{Prospective real-time correction for arbitrary
  head motion using active markers},'' \emph{\BIBforeignlanguage{eng}{Magnetic
  resonance in medicine: official journal of the Society of Magnetic Resonance
  in Medicine / Society of Magnetic Resonance in Medicine}}, vol.~62, no.~4,
  pp. 943--954, Oct. 2009.

\bibitem{ooi_echo-planar_2011}
M.~B. Ooi, S.~Krueger, J.~Muraskin, W.~J. Thomas, and T.~R. Brown,
  ``\BIBforeignlanguage{eng}{Echo-planar imaging with prospective
  slice-by-slice motion correction using active markers},''
  \emph{\BIBforeignlanguage{eng}{Magnetic resonance in medicine: official
  journal of the Society of Magnetic Resonance in Medicine / Society of
  Magnetic Resonance in Medicine}}, vol.~66, no.~1, pp. 73--81, Jul. 2011.

\bibitem{maintz_survey_1998}
J.~A. Maintz and M.~A. Viergever, ``A survey of medical image registration,''
  \emph{Medical image analysis}, vol.~2, no.~1, pp. 1--36, 1998.

\bibitem{friston_spatial_1995}
\BIBentryALTinterwordspacing
K.~J. Friston, J.~Ashburner, C.~D. Frith, J.-B. Poline, J.~D. Heather, and
  R.~S.~J. Frackowiak, ``\BIBforeignlanguage{en}{Spatial registration and
  normalization of images},'' \emph{\BIBforeignlanguage{en}{Human Brain
  Mapping}}, vol.~3, no.~3, pp. 165--189, Jan. 1995. [Online]. Available:
  \url{http://onlinelibrary.wiley.com/doi/10.1002/hbm.460030303/abstract}
\BIBentrySTDinterwordspacing

\bibitem{butts_interleaved_1994}
K.~Butts, S.~J. Riederer, R.~L. Ehman, R.~M. Thompson, and C.~R. Jack,
  ``Interleaved echo planar imaging on a standard {MRI} system,''
  \emph{Magnetic resonance in medicine}, vol.~31, no.~1, pp. 67--72, 1994.

\bibitem{kim_motion_1999}
B.~Kim, J.~L. Boes, P.~H. Bland, T.~L. Chenevert, and C.~R. Meyer,
  ``\BIBforeignlanguage{eng}{Motion correction in {fMRI} via registration of
  individual slices into an anatomical volume},''
  \emph{\BIBforeignlanguage{eng}{Magnetic resonance in medicine: official
  journal of the Society of Magnetic Resonance in Medicine / Society of
  Magnetic Resonance in Medicine}}, vol.~41, no.~5, pp. 964--972, May 1999.

\bibitem{schmitt_echo-planar_1998}
F.~Schmitt, M.~K. Stehling, and R.~Turner, \emph{Echo-planar imaging}.\hskip
  1em plus 0.5em minus 0.4em\relax Springer Science \& Business Media, 1998.

\bibitem{kim_intersection_2010}
K.~Kim, P.~Habas, F.~Rousseau, O.~Glenn, A.~J. Barkovich, C.~Studholme, and
  {others}, ``Intersection based motion correction of multislice {MRI} for
  3-{D} in utero fetal brain image formation,'' \emph{Medical Imaging, IEEE
  Transactions on}, vol.~29, no.~1, pp. 146--158, 2010.

\bibitem{kainz_fast_2015}
B.~Kainz, M.~Steinberger, W.~Wein, M.~Murgasova, C.~Malamateniou, K.~Keraudren,
  P.~Aljabar, M.~Rutherford, J.~Hajnal, and D.~Rueckert, ``Fast {Volume}
  {Reconstruction} from {Motion} {Corrupted} {Stacks} of 2d {Slices},'' 2015.

\bibitem{kotecha_gaussian_2003}
J.~H. Kotecha and P.~Djuric, ``Gaussian particle filtering,'' \emph{IEEE
  Transactions on Signal Processing}, vol.~51, no.~10, pp. 2592--2601, Oct.
  2003.

\bibitem{hill_medical_2001}
D.~L. Hill, P.~G. Batchelor, M.~Holden, and D.~J. Hawkes, ``Medical image
  registration,'' \emph{Physics in medicine and biology}, vol.~46, no.~3,
  p.~R1, 2001.

\bibitem{rueckert_nonrigid_1999}
D.~Rueckert, L.~I. Sonoda, C.~Hayes, D.~L.~G. Hill, M.~O. Leach, and D.~Hawkes,
  ``Nonrigid registration using free-form deformations: application to breast
  {MR} images,'' \emph{IEEE Transactions on Medical Imaging}, vol.~18, no.~8,
  pp. 712--721, Aug. 1999.

\bibitem{meyer_demonstration_1997}
C.~R. Meyer, J.~L. Boes, B.~Kim, P.~H. Bland, K.~R. Zasadny, P.~V. Kison,
  K.~Koral, K.~A. Frey, and R.~L. Wahl, ``Demonstration of accuracy and
  clinical versatility of mutual information for automatic multimodality image
  fusion using affine and thin-plate spline warped geometric deformations,''
  \emph{Medical image analysis}, vol.~1, no.~3, pp. 195--206, 1997.

\bibitem{rohr_landmark-based_2001}
K.~Rohr, H.~S. Stiehl, R.~Sprengel, T.~M. Buzug, J.~Weese, and M.~Kuhn,
  ``Landmark-based elastic registration using approximating thin-plate
  splines,'' \emph{Medical Imaging, IEEE Transactions on}, vol.~20, no.~6, pp.
  526--534, 2001.

\bibitem{maes_multimodality_1997}
F.~Maes, A.~Collignon, D.~Vandermeulen, G.~Marchal, and P.~Suetens,
  ``Multimodality image registration by maximization of mutual information,''
  \emph{Medical Imaging, IEEE Transactions on}, vol.~16, no.~2, pp. 187--198,
  1997.

\bibitem{mcrobbie_mri_2006}
D.~W. McRobbie, E.~A. Moore, M.~J. Graves, and M.~R. Prince, \emph{{MRI} from
  {Picture} to {Proton}}.\hskip 1em plus 0.5em minus 0.4em\relax Cambridge
  university press, 2006.

\bibitem{durbin_time_2012}
J.~Durbin and S.~J. Koopman, \emph{Time series analysis by state space
  methods}.\hskip 1em plus 0.5em minus 0.4em\relax Oxford University Press,
  2012, no.~38.

\bibitem{han_visual_2009}
B.~Han, Y.~Zhu, D.~Comaniciu, and L.~S. Davis,
  ``\BIBforeignlanguage{eng}{Visual tracking by continuous density propagation
  in sequential bayesian filtering framework},''
  \emph{\BIBforeignlanguage{eng}{IEEE transactions on pattern analysis and
  machine intelligence}}, vol.~31, no.~5, pp. 919--930, May 2009.

\bibitem{kalman_new_1960}
\BIBentryALTinterwordspacing
R.~E. Kalman, ``A {New} {Approach} to {Linear} {Filtering} and {Prediction}
  {Problems},'' \emph{Journal of Fluids Engineering}, vol.~82, no.~1, pp.
  35--45, Mar. 1960. [Online]. Available:
  \url{http://dx.doi.org/10.1115/1.3662552}
\BIBentrySTDinterwordspacing

\bibitem{julier_new_1997}
\BIBentryALTinterwordspacing
S.~J. Julier and J.~K. Uhlmann, ``New extension of the {Kalman} filter to
  nonlinear systems,'' vol. 3068, 1997, pp. 182--193. [Online]. Available:
  \url{http://dx.doi.org/10.1117/12.280797}
\BIBentrySTDinterwordspacing

\bibitem{wan_unscented_2000}
E.~Wan and R.~Van~der Merwe, ``The unscented {Kalman} filter for nonlinear
  estimation,'' in \emph{Adaptive {Systems} for {Signal} {Processing},
  {Communications}, and {Control} {Symposium} 2000. {AS}-{SPCC}. {The} {IEEE}
  2000}, 2000, pp. 153--158.

\bibitem{doucet_sequential_2000}
\BIBentryALTinterwordspacing
A.~Doucet, S.~Godsill, and C.~Andrieu, ``\BIBforeignlanguage{en}{On sequential
  {Monte} {Carlo} sampling methods for {Bayesian} filtering},''
  \emph{\BIBforeignlanguage{en}{Statistics and Computing}}, vol.~10, no.~3, pp.
  197--208, Jul. 2000. [Online]. Available:
  \url{http://link.springer.com/article/10.1023/A%3A1008935410038}
\BIBentrySTDinterwordspacing

\bibitem{nelder_simplex_1965}
\BIBentryALTinterwordspacing
J.~A. Nelder and R.~Mead, ``\BIBforeignlanguage{en}{A {Simplex} {Method} for
  {Function} {Minimization}},'' \emph{\BIBforeignlanguage{en}{The Computer
  Journal}}, vol.~7, no.~4, pp. 308--313, Jan. 1965. [Online]. Available:
  \url{http://comjnl.oxfordjournals.org/content/7/4/308}
\BIBentrySTDinterwordspacing

\bibitem{studholme_overlap_1999}
\BIBentryALTinterwordspacing
C.~Studholme, D.~L.~G. Hill, and D.~J. Hawkes, ``An overlap invariant entropy
  measure of 3d medical image alignment,'' \emph{Pattern Recognition}, vol.~32,
  no.~1, pp. 71 -- 86, 1999. [Online]. Available:
  \url{http://www.sciencedirect.com/science/article/pii/S0031320398000910}
\BIBentrySTDinterwordspacing

\bibitem{klein_automatic_2008}
S.~Klein, U.~A. van~der Heide, I.~M. Lips, M.~van Vulpen, M.~Staring, and J.~P.
  Pluim, ``Automatic segmentation of the prostate in 3d {MR} images by atlas
  matching using localized mutual information,'' \emph{Medical physics},
  vol.~35, no.~4, pp. 1407--1417, 2008.

\bibitem{staring_registration_2009}
M.~Staring, U.~A. van~der Heide, S.~Klein, M.~A. Viergever, and J.~P. Pluim,
  ``Registration of cervical {MRI} using multifeature mutual information,''
  \emph{Medical Imaging, IEEE Transactions on}, vol.~28, no.~9, pp. 1412--1421,
  2009.

\bibitem{oliveira_medical_2014}
F.~P. Oliveira and J.~M.~R. Tavares, ``Medical image registration: a review,''
  \emph{Computer methods in biomechanics and biomedical engineering}, vol.~17,
  no.~2, pp. 73--93, 2014.

\bibitem{park_improved_2004}
\BIBentryALTinterwordspacing
H.~Park, C.~R. Meyer, and B.~Kim, ``\BIBforeignlanguage{en}{Improved {Motion}
  {Correction} in {fMRI} by {Joint} {Mapping} of {Slices} into an {Anatomical}
  {Volume}},'' in \emph{\BIBforeignlanguage{en}{Medical {Image} {Computing} and
  {Computer}-{Assisted} {Intervention} – {MICCAI} 2004}}, ser. Lecture
  {Notes} in {Computer} {Science}, C.~Barillot, D.~R. Haynor, and P.~Hellier,
  Eds.\hskip 1em plus 0.5em minus 0.4em\relax Springer Berlin Heidelberg, Jan.
  2004, no. 3217, pp. 745--751. [Online]. Available:
  \url{http://link.springer.com/chapter/10.1007/978-3-540-30136-3_91}
\BIBentrySTDinterwordspacing

\bibitem{cocosco_brainweb:_1997}
C.~Cocosco, V.~Kollokian, {Kwan}, and A.~Evans, ``{BrainWeb}: {Online}
  {Interface} to a 3d {MRI} {Simulated} {Brain} {Database},''
  \emph{NeuroImage}, vol.~5, no.~4, 1997.

\bibitem{kim_comprehensive_2008}
\BIBentryALTinterwordspacing
B.~Kim, D.~T.~B. Yeo, and R.~Bhagalia, ``Comprehensive mathematical simulation
  of functional magnetic resonance imaging time series including motion-related
  image distortion and spin saturation effect,'' \emph{Magnetic Resonance
  Imaging}, vol.~26, no.~2, pp. 147--159, Feb. 2008. [Online]. Available:
  \url{http://www.sciencedirect.com/science/article/pii/S0730725X07003062}
\BIBentrySTDinterwordspacing

\bibitem{nichols_nonparametric_2002}
T.~E. Nichols and A.~P. Holmes, ``\BIBforeignlanguage{eng}{Nonparametric
  permutation tests for functional neuroimaging: a primer with examples},''
  \emph{\BIBforeignlanguage{eng}{Human brain mapping}}, vol.~15, no.~1, pp.
  1--25, Jan. 2002.

\bibitem{noll_estimating_1997}
D.~C. Noll, C.~R. Genovese, L.~E. Nystrom, A.~L. Vazquez, S.~D. Forman, W.~F.
  Eddy, and J.~D. Cohen, ``Estimating test-retest reliability in functional
  {MR} imaging {II}: application to motor and cognitive activation studies,''
  \emph{Magnetic Resonance in Medicine}, vol.~38, no.~3, pp. 508--517, 1997.

\bibitem{genovese_estimating_1997}
C.~R. Genovese, D.~C. Noll, and W.~F. Eddy, ``Estimating test-retest
  reliability in functional {MR} imaging {I}: {Statistical} methodology,''
  \emph{Magnetic Resonance in Medicine}, vol.~38, no.~3, pp. 497--507, 1997.

\bibitem{hoaglin_performance_1986}
D.~C. Hoaglin, B.~Iglewicz, and J.~W. Tukey, ``Performance of some resistant
  rules for outlier labeling,'' \emph{Journal of the American Statistical
  Association}, vol.~81, no. 396, pp. 991--999, 1986.

\bibitem{beisteiner_finger_2001}
R.~Beisteiner, C.~Windischberger, R.~Lanzenberger, V.~Edward, R.~Cunnington,
  M.~Erdler, A.~Gartus, B.~Streibl, E.~Moser, and L.~Deecke, ``Finger
  somatotopy in human motor cortex,'' \emph{Neuroimage}, vol.~13, no.~6, pp.
  1016--1026, 2001.

\end{thebibliography}

% that's all folks
\end{document}